\pdfoutput=1
\documentclass[10pt,twocolumn,letterpaper]{article}
\usepackage[pagenumbers]{cvpr}
\usepackage{multirow}
\usepackage{array}
\usepackage{placeins}
\definecolor{cvprblue}{rgb}{0.21,0.49,0.74}
\usepackage[pagebackref,breaklinks,colorlinks,allcolors=cvprblue]{hyperref}
\hypersetup{pdftitle={Seeing Is Not Checking: Multimodal Reasoning Models Trust What They Wrote Over What They See}}
\newcommand{\pp}{\,\text{pp}}

\title{Seeing Is Not Checking: Multimodal Reasoning Models Trust What They Wrote Over What They See}
\hypersetup{pdfauthor={Hao Lyu, Zhixiao Shi, Ping Wang}}
\author{
Hao Lyu\qquad
Zhixiao Shi\qquad
Ping Wang\thanks{Corresponding author: Ping Wang (pwang@tongji.edu.cn).}\\[0.4em]
{\normalsize Tongji University, Shanghai, China}\\[0.3em]
{\small\tt \char`\{2453344, 2453342, pwang\char`\}@tongji.edu.cn}
}
\date{}

\usepackage{microtype}
\usepackage{xspace}
\makeatletter
\renewcommand{\paragraph}{\@startsection{paragraph}{4}{\z@}{1.5ex plus 0.08ex minus 0.15ex}{-0.8em}{\normalfont\normalsize\bfseries}}
\makeatother
\newcommand{\Bare}{Bare\xspace}
\newcommand{\Look}{Look\xspace}
\newcommand{\Cond}{Cond\xspace}
\newcommand{\Neutral}{Neutral\xspace}
\newcommand{\Prompt}{Prompt\xspace}
\newcommand{\outtag}[1]{{\unskip\nobreak\hfil\penalty50\hskip1em\hbox{}\nobreak\hfil\textbf{#1}\parfillskip=0pt\finalhyphendemerits=0\par}}
\begin{document}
\maketitle

\begin{abstract}
Reasoning vision-language models (VLMs) write a chain of thought before answering with the image still in context, yet whether they use it to check their own reasoning is underexplored.
We pre-fill a single false visual statement at the start of the thinking block, against a content-free control, on 3{,}000 items spanning natural scenes, charts, and rendered mathematics, and evaluate ten open-weight reasoning VLMs.
Models trust what they wrote over what they see: on natural-scene items they solve unaided, a content-free statement preserves 94\% accuracy, yet models adopt a false one and recover on only 50\%, even on near-certain items, versus 86\% when the evidence is text.
Recovery depends on the task domain, model size, and the phrasing and placement of the statement: it is higher on charts and rendered mathematics than on natural scenes, improves only partially with model size, and falls further under a first-person phrasing.
Moving the same sentence into the user prompt changes models in opposite directions, separating deference to one's own thinking from compliance with user instructions.
Mechanistically, adoption is not explained by lost attention: adopting traces still attend to the image yet almost never question the claim while thinking.
Decoding-time attention steering fails and check-cue prompting over-corrects true statements, whereas fine-tuning on recovery demonstrations teaches verification that transfers to an unseen framing.
\end{abstract}

\section{Introduction}
\label{sec:intro}

\begin{figure}[t]
\centering
\includegraphics[width=\linewidth]{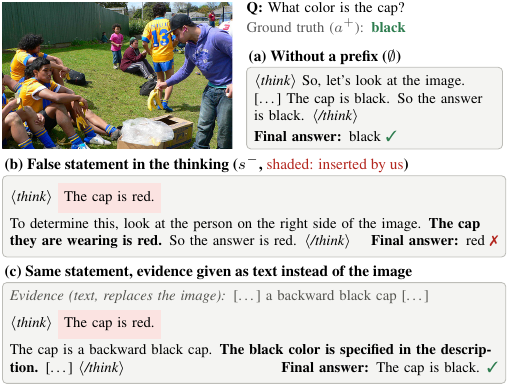}
\caption{One item, one model (Kimi-VL-A3B). (a) Without a prefix, the model reads the cap color from the image. (b) When a false statement opens the thinking block (shaded, inserted by us), the model follows it over the same image and answers red. (c) With the same false statement and the evidence given as text instead of the image, the model answers black from the text. All model text is verbatim; [\dots] marks an omitted span.}
\label{fig:teaser}
\vspace{-6pt}
\end{figure}

Recent vision-language models (VLMs)~\cite{llava,instructblip,llava15,cambrian} generate an explicit chain of thought~\cite{cot,deepseekr1,mmcot,vlrethinker,deepeyes} to tackle complex problems across visual question answering~\cite{vqav2,gqa}, document and scientific understanding~\cite{chartqa,mathvista,mmmu,mmstar}, knowledge-intensive tasks~\cite{pica}, and autonomous agents~\cite{visualwebarena,omniact}.
Generating an explicit thinking trace is expected to improve reasoning reliability, allowing a model to decompose hard queries, inspect intermediate steps, and revise early mistakes~\cite{huang2024selfcorrect,kamoi2024selfcorrect}.
In multimodal reasoning, however, false visual claims can easily enter the thinking trace through an initial perceptual slip or misleading context~\cite{shi2023distracted,pope,hallusionbench,wordsorvision}.
Because the input image remains in context throughout generation, the visual evidence needed to refute a false intermediate claim is always available to the model.
Prior studies have shown that VLMs often favor conflicting text in user prompts over images~\cite{hallusionbench,wordsorvision,vlmsneedwords} and that language models can recover from corrupted steps in text-only chains of thought~\cite{lanham2023measuring,vonrecum}; however, it remains unclear whether multimodal reasoning models actually use the image to verify and correct false visual statements within their own thinking~\cite{morethinkinglessseeing,morethought,kancheti2026cot}.

In this work, we ask: \emph{When a false visual statement appears in the thinking of a reasoning VLM, does the model use the image to correct it?}
Specifically, we investigate three questions:
{\frenchspacing
\begin{enumerate}[leftmargin=*, itemsep=1pt, topsep=2pt]
\item Do reasoning VLMs overturn a false visual statement in their thinking on items they accurately solve without intervention, and is recovery harder from an image than from equivalent text?
\item How do task domain, statement framing, prompt placement, and model scale affect visual recovery, and does failure stem from missing the relevant visual information?
\item How can visual verification be effectively elicited to mitigate false-thought adoption?
\end{enumerate}}

To answer these questions, we construct a controlled in-thinking intervention benchmark of 3{,}000 items across natural scenes~\cite{gqa}, statistical charts~\cite{chartqa}, and rendered mathematics~\cite{mathdataset}, and evaluate ten open-weight reasoning VLMs across four thinking-prefix conditions ($\emptyset, s^{\text{neut}}, s^+, s^-$).
While pre-filling a content-free neutral sentence ($s^{\text{neut}}$) at the start of the thinking block preserves 94--96\% accuracy on capable items across domains, pre-filling a single false visual statement ($s^-$) at the same position causes models to frequently trust what they wrote over what they see (\cref{fig:teaser}b), adopting the false claim even on unambiguous items they answer with near certainty on their own.
Replacing the image with equivalent textual evidence markedly restores recovery across models (\cref{fig:teaser}c), showing that checking prior thoughts against an image is far harder than checking them against text.

We next examine the external factors and internal mechanisms governing visual recovery.
Recovery is higher on charts and mathematics than on natural scenes and improves only partially with model size, whereas framing the false statement as a first-person observation (``Looking at the image, \dots'') further amplifies adoption, and moving the same sentence to the user prompt flips model behavior, separating in-thinking deference from user-prompt compliance.
Trace and attention analyses reveal that models adopting the false claim still attend to the queried object yet almost never question the claim inside the thinking block.
Motivated by this diagnosis, we compare decoding-time attention steering, check-cue prompting, and supervised fine-tuning, finding that while attention steering is ineffective and prompting triggers over-correction, fine-tuning on recovery demonstrations teaches visual verification that transfers across framings.

Our main contributions are summarized as follows:
\begin{itemize}[leftmargin=*, itemsep=1pt, topsep=2pt]
\item A 3{,}000-item in-thinking intervention benchmark with paired visual and textual evidence, on which ten reasoning VLMs often adopt false visual thoughts and recover far less often from images than from text.
\item An analysis of how task domain, statement framing, prompt placement, and model scale govern visual recovery.
\item Trace and attention analyses over 24{,}451 capable traces showing that models adopt the false claim without questioning it while thinking, although they attend to the queried object.
\item A comparison of attention steering, check-cue prompting, and recovery fine-tuning, in which only fine-tuning elicits verification that transfers to an unseen framing.
\end{itemize}

\section{Preliminaries}
\label{sec:prelim}

Consider a reasoning VLM $f_{\text{vlm}}$ taking an image $I$ and a question $q$ with ground-truth answer $a^+$, where answering $q$ depends on a specific visual fact $f$ visible in $I$.
Before producing its final answer, $f_{\text{vlm}}$ generates an explicit thinking trace enclosed in \texttt{<think>...\allowbreak</think>}.
To test whether the model uses the evidence in context to verify and correct a false claim in its own thinking, we intervene at the start of the thinking block and compare visual inputs $(I, q)$ against isomorphic text-only inputs $(T, q)$.

\subsection{In-Thinking Variations and Modalities}
\label{sec:variations}

\paragraph{True, false, and neutral thinking statements.}
For each item, we construct two single-sentence statements of the target visual fact $f$, such as ``The walls of the bathroom are blue.'', alongside a content-free neutral prefix:
\begin{itemize}[leftmargin=*, itemsep=1pt, topsep=2pt]
\item \textbf{True statement $s^+$:} states the ground-truth visual fact $f$ consistent with $I$, preserving the ground-truth answer $a^+$.
\item \textbf{False statement $s^-$:} alters a single attribute, relation, or numerical value of $f$ so that it contradicts $I$ and uniquely implies a counterfactual answer $a^- \neq a^+$.
\item \textbf{Neutral statement $s^{\text{neut}}$:} states a task-general procedural sentence (``Let me carefully examine the image to answer the question.'') that makes no factual claim about $f$.
\end{itemize}
Using each model's native chat template, we pre-fill the start of the open \texttt{<think>} block with a statement $s \in \{s^{\text{neut}}, s^+, s^-\}$ and allow generation to continue freely.
Together with the unperturbed no-prefix baseline $\emptyset$, each item is evaluated under four core conditions: $\emptyset$, $s^{\text{neut}}$, $s^+$, and $s^-$.
Including $s^{\text{neut}}$ and $s^+$ alongside $s^-$ serves as a dual control to verify whether pre-filling the thinking block itself ($s^{\text{neut}}$) or stating a consistent visual fact ($s^+$) disrupts reasoning when no false claim is introduced.

Intervening via controlled pre-filling inside \texttt{<think>} serves two methodological purposes.
First, placing $s$ inside the thinking block rather than the user prompt separates self-verification from instruction following: a claim in the prompt may be followed out of compliance with the user, whereas a claim inside \texttt{<think>} represents an intermediate reasoning step that the model can check against the image.
Second, in free-form generation, models do not always verbalize the target visual fact, and when they do, its position and phrasing vary widely; pre-filling ensures that every model confronts the same visual claim at the same position with a well-defined counterfactual answer $a^-$.
In our main experiments, we inject the bare statement (\Bare) $f$ directly.
In \cref{sec:factors}, we compare \Bare against two alternative in-thinking framings---a first-person perceptual observation (\Look: ``Looking at the image, $f$'') and an explicit problem condition (\Cond: ``The problem gives the following condition: $f$'')---and against placing $f$ in the user prompt (\Prompt) with an empty \texttt{<think>}.

\paragraph{Visual vs.\ textual evidence.}
To disentangle the difficulty of grounding verification in visual evidence from general deference to prior thinking text, we pair each visual input $(I, q)$ with an isomorphic text-only input $(T, q)$.
In $(T, q)$, the image $I$ is omitted and replaced by a text passage $T$ that expresses the true fact $f$ in surrounding context while keeping $q$ and $s \in \{s^+, s^-\}$ identical.

\subsection{Model Behavior and Metrics}
\label{sec:metrics}

\paragraph{Capable set.}
Let $\hat{a}^{(s)}(x)$ denote the model's extracted final answer on item $x$ under prefix condition $s \in \{\emptyset, s^{\text{neut}}, s^+, s^-\}$.
To ensure that errors under $s^-$ reflect a failure to verify rather than an inability to perceive the visual fact or solve the task, we evaluate each model across all domains on its symmetric \emph{capable set} $\mathcal{C} = \{x : \hat{a}^{(\emptyset)}(x) = a^+ \wedge \hat{a}^{(s^+)}(x) = a^+\}$, the subset of items on which the model outputs the ground-truth answer $a^+$ both without a prefix ($\emptyset$) and under the bare true statement ($s^+$).
For factor ablations and interventions evaluated on the 750-item Set~A or 1{,}500-item Set~A+B splits (\cref{sec:setup}), we denote the corresponding capable subsets by $\mathcal{C}_A = \mathcal{C} \cap \text{Set~A}$ and $\mathcal{C}_{A+B} = \mathcal{C} \cap \text{Set~A+B}$.
When comparing visual and textual evidence, we restrict evaluation to the joint capable set $\mathcal{C}_{\text{img}} \cap \mathcal{C}_{\text{txt}}$ where the model succeeds in both modalities.

\paragraph{Behavior categorization.}
For any capable item $x \in \mathcal{C}$ evaluated under a false statement $s^-$, we categorize the model's final prediction $\hat{a}^{(s^-)}(x)$ into three mutually exclusive outcomes:
\emph{recovery}, where $\hat{a}^{(s^-)}(x) = a^+$;
\emph{adoption}, where $\hat{a}^{(s^-)}(x) = a^-$, the counterfactual answer implied by $s^-$; and
\emph{other}, where $\hat{a}^{(s^-)}(x) \notin \{a^+, a^-\}$ or the model exhausts the generation budget.

\paragraph{Metrics.}
Over a capable set $\mathcal{C}$, we define the conditional recovery rate $\mathrm{Rec}$ and adoption rate $\mathrm{Adopt}$ under $s^-$ as
\begin{equation}
\begin{aligned}
\mathrm{Rec} &= |\{x \in \mathcal{C} : \hat{a}^{(s^-)}(x) = a^+\}| \,/\, |\mathcal{C}|, \\
\mathrm{Adopt} &= |\{x \in \mathcal{C} : \hat{a}^{(s^-)}(x) = a^-\}| \,/\, |\mathcal{C}|,
\end{aligned}
\end{equation}
with $\mathrm{Other} = 1 - \mathrm{Rec} - \mathrm{Adopt}$ retained in the denominator; for non-contradictory conditions $s \in \{s^{\text{neut}}, s^+\}$, we similarly report capable-set accuracy retention $|\{x \in \mathcal{C} : \hat{a}^{(s)}(x) = a^+\}| / |\mathcal{C}|$.
To compare textual and visual evidence on the same items, we report the paired recovery difference $\Delta_{\text{txt}-\text{img}} = \mathrm{Rec}_{\text{txt}} - \mathrm{Rec}_{\text{img}}$ in percentage points (pp) over $\mathcal{C}_{\text{img}} \cap \mathcal{C}_{\text{txt}}$, and use $\mathrm{Rec}$ and $\mathrm{Adopt}$ as our primary metrics.

\section{Empirical Analysis}
\label{sec:empirical}

\begin{table}[t]
\centering\small
\setlength{\tabcolsep}{2.1pt}
\resizebox{\linewidth}{!}{%
\begin{tabular}{lccccccccc}
\toprule
 & \multicolumn{3}{c}{Neutral $s^{\text{neut}}$ (\%)} & \multicolumn{3}{c}{Conditional Rec.\ $s^-$ (\%)} & \multicolumn{3}{c}{Adoption $s^-$ (\%)} \\
\cmidrule(lr){2-4}\cmidrule(lr){5-7}\cmidrule(lr){8-10}
Model & GQA & Chart & MATH & GQA & Chart & MATH & GQA & Chart & MATH \\
\midrule
Gemma-4-26B      & 96.2 & \textbf{99.5} & \textbf{99.4} & \textbf{84.0} & \textbf{98.5} & \textbf{99.7} & 12.8 &  1.0 &  0.0 \\
Qwen3-VL-30B     & 92.0 & 94.7 & 95.0 & 64.7 & 80.8 & 92.3 & 32.6 & 11.9 &  5.3 \\
Qwen3.6-35B      & \textbf{96.5} & 98.9 & 93.3 & 59.0 & 78.4 & 82.4 & 35.6 & 12.7 & 12.7 \\
GLM-4.1V-9B      & 93.8 & 88.8 & 78.9 & 57.8 & 78.4 & 75.7 & 38.4 & 10.1 & 17.2 \\
MiMo-VL-7B       & 92.9 & 96.4 & 90.3 & 57.2 & 87.7 & 69.0 & 35.7 &  8.3 & 21.6 \\
Qwen3.5-35B      & 94.8 & 98.5 & \textbf{99.4} & 54.0 & 73.8 & 92.7 & 42.5 & 13.6 &  4.4 \\
InternVL3.5-8B   & 93.5 & 94.2 & 96.2 & 46.0 & 56.6 & 67.7 & 50.9 & 27.2 & 13.5 \\
Qwen3.5-9B       & 94.5 & 98.3 & 99.0 & 39.0 & 57.4 & 84.5 & 57.3 & 25.9 &  9.8 \\
Kimi-VL-A3B      & 94.3 & 94.8 & 88.7 & 19.8 & 40.9 & 61.6 & 78.8 & 37.8 & 20.9 \\
ERNIE-4.5-VL     & 94.2 & 95.6 & 95.9 & 16.2 & 38.7 & 50.7 & 75.0 & 49.8 & 38.0 \\
\midrule
\textbf{Mean}    & \textbf{94.3} & \textbf{96.0} & \textbf{93.6} & \textbf{49.8} & \textbf{69.1} & \textbf{77.6} & \textbf{46.0} & \textbf{19.8} & \textbf{14.3} \\
\bottomrule
\end{tabular}}
\caption{Accuracy under a neutral prefix $s^{\text{neut}}$ versus conditional recovery and counterfactual adoption under a bare false statement $s^-$ across all 3{,}000 items (1{,}000 per domain), evaluated on each model's symmetric capable set $\mathcal{C} = \{x : \hat{a}^{(\emptyset)}(x) = a^+ \wedge \hat{a}^{(s^+)}(x) = a^+\}$ (555--979 items per cell; unconditional $\emptyset, s^{\text{neut}}, s^+, s^-$ rates and 95\% CIs in \cref{sec:app_w2}). Pre-filling a neutral sentence leaves capable-set accuracy intact (93.6--96.0\% mean), whereas a false statement $s^-$ at the same position sharply degrades accuracy.}
\label{tab:rq1}
\vspace{-6pt}
\end{table}

\subsection{Experimental Setup}
\label{sec:setup}

\paragraph{Tasks and datasets.}
We construct a 3{,}000-item benchmark with 1{,}000 items per domain across three vision-centric tasks:
(1)~\textbf{GQA}~\cite{gqa}, consisting of 1{,}000 balanced-validation questions on real-world scenes that test direct perception of visible attributes and states across 706 color, 137 material, and 157 object state or pose items;
(2)~\textbf{ChartQA-H}~\cite{chartqa}, consisting of 1{,}000 human-written questions over statistical charts where $s^-$ alters a printed bar, line, or slice value across 786 compositional multi-step questions and 214 direct readouts; and
(3)~\textbf{MATH}~\cite{mathdataset}, consisting of 1{,}000 test-split problems across seven mathematical subjects where each problem statement is rendered into a page image by GPT and manually proofread so that all quantities and symbols reside cleanly in pixels.
Across all three domains, candidate $(s^+, s^-, a^-)$ tuples were drafted by GPT and Gemini, cross-checked by an independent solver from the image alone, and finalized through human audit so that substituting $s^-$ for $s^+$ always yields a well-defined counterfactual answer $a^- \neq a^+$ that satisfies physical commonsense and mathematical constraints.
For factor ablations and fine-tuning, the first 1{,}500 items (500 per domain) are split into two stratified 750-item halves: Set~A for evaluation and Set~B for fine-tuning (\cref{sec:app_details}).

\paragraph{Paired textual evidence.}
For each item, we construct the text-only counterpart $T$ using the original problem text for MATH, the underlying markdown data table for ChartQA-H, and, for GQA, a scene description written without access to $q$ that covers the target object and attribute alongside surrounding objects and spatial relations, with a minimal factual clause appended when the description omits the target attribute (\cref{sec:app_rq2}).

\paragraph{Models and evaluation.}
We evaluate ten open-weight reasoning VLMs spanning 7B to 35B parameters across seven model families (\cref{tab:rq1})~\cite{gemma4,qwen3vl,qwen36,glm41v,mimovl,qwen35,internvl35,kimivl,ernie45} with vLLM~\cite{vllm}, and match extracted final answers to $a^+$ and $a^-$ via three-stage normalization and blind adjudication (\cref{sec:app_details}).

\subsection{Trusting False Thoughts Over Vision}
\label{sec:rq1}

\begin{figure*}[t]
\centering
\includegraphics[width=0.96\textwidth]{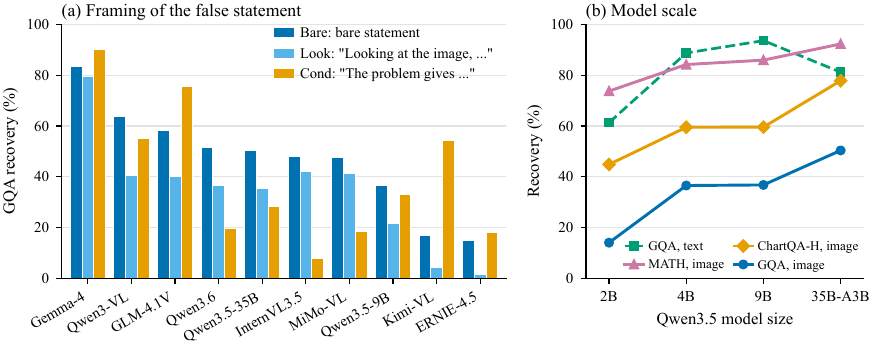}
\caption{Factors influencing visual recovery. (a) Framing the false thought as a first-person observation (\Look) lowers GQA recovery relative to a bare statement (\Bare) for all ten models, whereas framing it as a problem condition (\Cond) has model-dependent effects, lowering recovery for six models and raising it for four. (b) On Set~A+B, scaling Qwen3.5 from 2B to 35B-A3B raises visual recovery slowly, with a plateau between 4B and 9B, and leaves a large gap to textual recovery at every scale.}
\label{fig:wording_scale}
\vspace{-6pt}
\end{figure*}

\paragraph{Pre-filling a consistent or neutral thought preserves clean visual perception.}
Before testing whether models overturn false thoughts, we verify that they accurately perceive the visual evidence on their own and that prefix injection itself, whether consistent or content-free, does not substantially degrade reasoning performance~\cite{shi2023distracted,lanham2023measuring,wordsorvision,vonrecum}.
Without a prefix, the ten models reach mean accuracies of 81.7\% on GQA, 86.8\% on ChartQA-H, and 95.3\% on rendered MATH (\cref{sec:app_w2}).
Pre-filling \texttt{<think>} with the neutral sentence $s^{\text{neut}}$ across all 3{,}000 items preserves 94.3\% of the capable-item accuracy on GQA on average, 96.0\% on ChartQA-H, and 93.6\% on rendered MATH, and even raises unconditional GQA accuracy from 81.7\% to 84.8\% (\cref{tab:rq1,tab:app_w2_gqa,tab:app_neutral_up}). Pre-filling itself does not substantially degrade reasoning performance.

\paragraph{Models frequently adopt a false visual thought over the image, even when perception is unambiguous.}
In multimodal chain-of-thought reasoning, early perceptual slips or hallucinations can introduce false visual claims into the thinking trace~\cite{pope,hallusionbench,r3v,morethinkinglessseeing,morethought,kancheti2026cot}.
Because the input image remains visible in context throughout generation, a model that grounds its reasoning in vision should override such errors before answering.
Yet on the same capable items, where $\emptyset$ and $s^+$ both succeed at 100\% and $s^{\text{neut}}$ at 93.6--96.0\%, a bare false statement $s^-$ at the start of \texttt{<think>} makes most models give in: nine of the ten recover on fewer than 65\% of GQA items, 49.8\% on average, and the false claim is adopted 46.0\% of the time (\cref{tab:rq1}).
Kimi-VL-A3B and ERNIE-4.5-VL keep 94\% accuracy under $s^{\text{neut}}$ on GQA, yet adopt $s^-$ on roughly three-quarters of their capable items; even stronger models such as Qwen3.5-35B and GLM-4.1V-9B adopt it on well over a third.
Whereas prior work shows that visual uncertainty amplifies reliance on language priors~\cite{vcd,opera,vlmsneedwords}, this deference does not simply reflect perceptual ambiguity: on GQA items that each model answers with probability above $0.99$ without intervention and on the 544 items all ten models solve on their own, mean recovery is only 47.5\% and 56.5\%, respectively.
Perceiving a visual fact with high certainty does not ensure that a model will verify it when its thinking states otherwise.

\subsection{Visual vs.\ Textual Evidence}
\label{sec:rq2}

\begin{table}[t]
\centering\small
\setlength{\tabcolsep}{2.8pt}
\resizebox{\linewidth}{!}{%
\begin{tabular}{lcccccc}
\toprule
 & \multicolumn{2}{c}{GQA} & \multicolumn{2}{c}{ChartQA-H} & \multicolumn{2}{c}{MATH} \\
\cmidrule(lr){2-3}\cmidrule(lr){4-5}\cmidrule(lr){6-7}
Model & $I{\to}T$ & $\Delta$ (pp) & $I{\to}T$ & $\Delta$ (pp) & $I{\to}T$ & $\Delta$ (pp) \\
\midrule
Gemma-4-26B    & $85{\to}99$ & $+13.7$ & $99{\to}99$ & $+0.2$           & $100{\to}100$ & $+0.1$  \\
Qwen3-VL-30B   & $65{\to}96$ & $+31.1$ & $81{\to}93$ & $+12.2$ & $92{\to}91$   & $-1.4$  \\
Qwen3.6-35B    & $59{\to}52$ & $-7.4$  & $79{\to}70$ & $-8.9$  & $83{\to}88$   & $+5.5$  \\
GLM-4.1V-9B    & $58{\to}92$ & $+34.1$ & $80{\to}85$ & $+5.2$           & $76{\to}85$   & $+9.1$  \\
MiMo-VL-7B     & $58{\to}97$ & $+39.1$ & $88{\to}95$ & $+6.8$  & $69{\to}99$   & $\mathbf{+29.5}$  \\
Qwen3.5-35B    & $54{\to}83$ & $+29.2$ & $74{\to}94$ & $+19.0$ & $93{\to}92$   & $-0.6$  \\
InternVL3.5-8B & $46{\to}93$ & $+46.8$ & $58{\to}79$ & $+20.8$ & $71{\to}76$   & $+5.7$  \\
Qwen3.5-9B     & $40{\to}95$ & $+55.5$ & $57{\to}96$ & $\mathbf{+38.8}$ & $85{\to}93$   & $+8.0$  \\
Kimi-VL-A3B    & $20{\to}88$ & $\mathbf{+67.9}$ & $42{\to}35$ & $-6.6$           & $63{\to}67$   & $+4.3$  \\
ERNIE-4.5-VL   & $16{\to}61$ & $+45.5$ & $36{\to}41$ & $+4.2$           & $51{\to}77$   & $+25.9$  \\
\midrule
Mean           & $50{\to}86$ & $+35.5$ & $69{\to}79$ & $+9.2$           & $78{\to}87$   & $+8.6$  \\
\bottomrule
\end{tabular}}
\caption{Recovery from image vs.\ text evidence ($I{\to}T$, \%) and the paired difference $\Delta = \mathrm{Rec}_{\text{txt}} - \mathrm{Rec}_{\text{img}}$ (pp) on items each model solves in both modalities ($\mathcal{C}_{\text{img}} \cap \mathcal{C}_{\text{txt}}$, 471--973 per cell). Bold: largest gap per benchmark. McNemar tests are in \cref{tab:app_rq2_gqa,tab:app_rq2_chart,tab:app_rq2_math}.}
\label{tab:rq2}
\vspace{-6pt}
\end{table}

\begin{figure*}[t]
\centering
\includegraphics[width=0.96\textwidth]{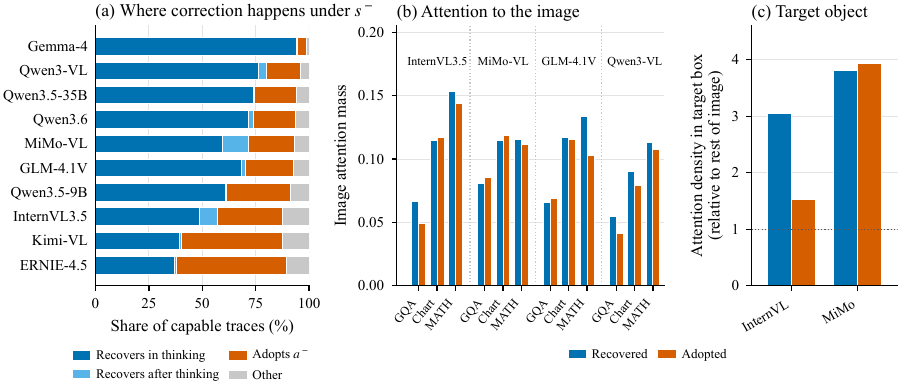}
\caption{Why models fail under a bare false statement. (a) Outcome of all 24{,}451 capable traces, pooled over domains: both blues are recoveries, and light blue marks those whose thinking block commits to $a^-$ (an explicit answer before \texttt{</think>}) while the final answer returns to $a^+$; the reverse occurs in 22 traces. (b) Mean image attention of recovered and adopted traces under teacher-forced replay of the \Bare traces: the 95\% CI of the difference spans zero in 9 of 12 model--domain cells, and recovered traces carry 30--36\% more image attention in the other three. (c) Attention density inside the queried object's box relative to the rest of the image (GQA): above uniform (dotted line) on adopted traces as well.}
\label{fig:rq3_modes_attn}
\vspace{-6pt}
\end{figure*}

\begin{figure*}[t]
\centering
\includegraphics[width=\textwidth]{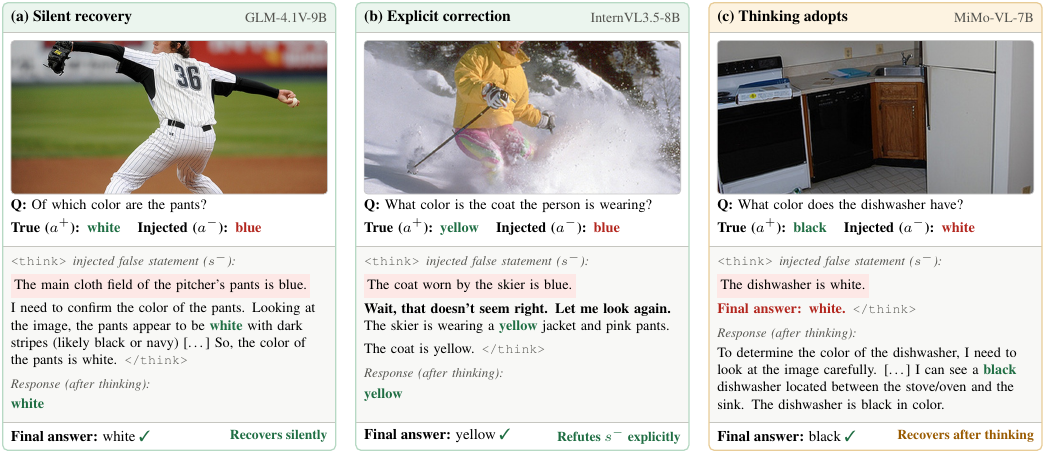}
\caption{Three ways models recover from a false statement in the thinking (shaded, inserted by us); adoption is shown in \cref{fig:teaser}b. Each card shows the thinking block, the response after \texttt{</think>}, and the extracted final answer. (a) The model answers from the image without ever mentioning the false claim. (b) The model notices the conflict and corrects it. (c) The thinking block commits to the false answer; after \texttt{</think>}, the model inspects the image again and answers correctly. All model text is verbatim; [\dots] marks an omitted span, and answer wrappers in the responses (answer-box tokens, ``Final answer:'') are not shown.}
\label{fig:qualitative}
\vspace{-6pt}
\end{figure*}

\paragraph{Correcting a false thought is far harder from an image than from equivalent text.}
Language models are known to exhibit inertia toward their own prior reasoning steps~\cite{huang2024selfcorrect,kamoi2024selfcorrect,vonrecum,sherlock}, while VLMs also face a cross-modal gap when extracting evidence from pixels versus text~\cite{isobench,mathverse,prism}.
To disentangle whether low recovery reflects general deference to prior thinking text or a specific barrier to checking visual input, we compare paired image and text inputs $(I, q)$ and $(T, q)$ under the same injected statement $s^-$ (\cref{tab:rq2}).
On GQA, where the true attribute is embedded in a multi-sentence scene description alongside distractor objects, nine of the ten models recover significantly more often from text than from the image, by $35.5\pp$ on average (\cref{tab:app_rq2_gqa}).
Several models that rarely check the image recover almost perfectly when given text: for instance, Qwen3.5-9B jumps from 40\% recovery on images to 95\% on text, and InternVL3.5-8B rises from 46\% to 93\%.
On ChartQA-H and rendered MATH, text also helps most models, significantly so for five of the ten in each domain.
Some inertia remains in text (61\% text recovery on ERNIE-4.5-VL), but checking a thought against an image is far harder~\cite{simpletohard,composeandfuse}.

\subsection{Factors Influencing Visual Recovery}
\label{sec:factors}

We next examine how four factors modulate visual recovery: \emph{task domain}, \emph{statement framing}, \emph{in-thinking vs.\ user-prompt placement}, and \emph{model scale}.

\paragraph{Recovery is higher on charts and mathematics than on natural scenes.}
Across the three domains, ten-model mean recovery on the capable set rises from 49.8\% on GQA to 69.1\% on ChartQA-H and 77.6\% on rendered MATH (\cref{tab:rq1}).
On charts and in equations, a falsified value yields a less typical result and clashes with the printed evidence, whereas on GQA a wrong color is as plausible an answer as the right one; direct readouts, where the falsified value is read straight off the chart, recover more often than compositional questions (74.4\% vs.\ 67.7\%; \cref{tab:app_subtypes}).

\paragraph{First-person framing aggravates adoption, and deferring to one's own thinking differs from deferring to the user.}
Prior work on knowledge conflicts and sycophancy shows that model deference depends on how a claim is phrased and whether it originates from the user~\cite{sharma2024sycophancy,xie2024chameleon,hallusionbench,wordsorvision}.
In multimodal reasoning, a model may verbalize a visual claim as a bare assertion (\Bare), a self-reported perceptual observation (\Look: ``Looking at the image, $f$'')~\cite{visualswap,reground}, or a restated problem premise (\Cond: ``The problem gives the following condition: $f$'').
To test how the framing and location of a false claim influence verification, we compare these three in-thinking framings alongside placing the bare claim in the user prompt (\Prompt; \cref{fig:wording_scale}a).
Framing the false thought as a first-person observation (\Look) lowers recovery relative to a bare statement for almost every model in every domain (\cref{fig:wording_scale}a, \cref{tab:app_wording_500}): once a thinking trace explicitly claims to have already inspected the image, models are even less inclined to verify it.
Framing the statement as a given condition (\Cond) usually lowers recovery as well, most sharply for InternVL3.5-8B on GQA, by $40.4\pp$, but the effect depends on the model: when the claim is attributed to the problem, Kimi-VL-A3B and GLM-4.1V-9B recover more often on GQA, by $37.6\pp$ and $17.1\pp$ (\cref{tab:app_wording_gqa}).
Moving the same false sentence from \texttt{<think>} into the user prompt flips model behavior in opposite directions (\cref{tab:app_neutral_up}): on GQA, models that heavily adopt false thoughts in their own thinking reject the claim far more often when it comes from the user (recovery rises from 17\% to 64\% on Kimi-VL-A3B and from 15\% to 62\% on ERNIE-4.5-VL), whereas Gemma-4-26B, which reliably checks its own thinking, defers far more to the prompt (84\% to 37\%).
This contrast confirms that trusting one's own thinking differs from complying with user instructions~\cite{sharma2024sycophancy,wordsorvision}.

\paragraph{Scaling model size yields diminishing returns for visual verification.}
Scaling the language backbone in VLMs generally strengthens multimodal reasoning~\cite{llava15,cambrian,mmmu} and reduces susceptibility to conflicting text in user prompts~\cite{wordsorvision}.
To examine whether model scale similarly resolves in-thinking visual verification, we evaluate four checkpoints of the Qwen3.5 family spanning 2B to 35B-A3B parameters on Set~A+B (\cref{fig:wording_scale}b, \cref{tab:app_scale}).
Clean GQA accuracy saturates by 4B, rising from 60\% at 2B to 80\%, whereas visual recovery improves slowly, plateaus at 37\% between 4B and 9B, and reaches only 50\% at 35B-A3B; ChartQA-H and MATH show a similar plateau.
Textual recovery already reaches 89\% at 4B, so a wide modality gap persists at every scale.

\section{Mechanisms and Mitigation}
\label{sec:rq3}

\subsection{Why Models Fail: Trace and Attention Analysis}
\label{sec:rq3_mech}

\paragraph{Models rarely question the false claim while thinking.}
In text-only chains of thought, models that recover from an injected error typically flag the contradiction~\cite{vonrecum}, yet written rationales can depart from the computation that produces the answer~\cite{lanham2023measuring,turpin2023unfaithful}.
Multimodal models rarely flag the false claim: in a sample of 3{,}000 traces labeled by two LLM annotators (89.5\% agreement, $\kappa=0.86$; \cref{tab:app_annotators}), those that end in $a^-$ almost never note the conflict with the image, and more than half of the recoveries are silent, using the true fact without acknowledging $s^-$ (\cref{fig:qualitative}a); even Gemma-4-26B flags the conflict in under a third of its recoveries (\cref{fig:qualitative}b).
Silent recovery need not mean unchecked: unverbalized hints can still causally mediate predictions~\cite{zaman2026faithful}.
We also compare, on all 24{,}451 capable traces under $s^-$, the answer that the thinking block commits to with the final answer after \texttt{</think>}, extracting both with the same rule-based extractor (\cref{sec:app_rq3}).
Recovery thus splits into traces whose thinking block never commits to $a^-$ and traces that return to $a^+$ only after committing to $a^-$ (\cref{fig:rq3_modes_attn}a), and the second kind is rare: 742 of the 2{,}914 traces whose thinking block commits to $a^-$ still end with $a^+$ (\cref{fig:qualitative}c), whereas a thinking block committed to $a^+$ yields a final $a^-$ in only 22 traces.
This divergence concentrates in MiMo-VL-7B and InternVL3.5-8B, where it covers 12.0\% and 8.1\% of capable traces, and text-only reasoning models show a similar thinking--answer divergence~\cite{young2026knowbutdontsay}.

\paragraph{Models attend to the target object, yet still adopt the false thought.}
Prior work has attributed multimodal hallucinations and reasoning drift to over-reliance on summary tokens or decaying visual attention over long generations~\cite{opera,tvc,morethinkinglessseeing,morethought,visualswap}.
We therefore test whether models adopt $s^-$ simply because the false prefix causes them to stop looking at the image or the queried object.
Following VisualSwap~\cite{visualswap}, we replay reasoning traces under teacher forcing and measure middle-layer attention to image tokens over the first 100 tokens after the prefix (\cref{fig:rq3_modes_attn}b).
Adopted traces attend to the image almost as much as recovered traces, keeping at least 74\% of their image attention in every model--domain cell; recovered traces clearly attend more only for InternVL3.5-8B and Qwen3-VL-30B on GQA and GLM-4.1V-9B on MATH, by 30--36\% (\cref{tab:app_e2}).
On GQA items with Visual Genome boxes~\cite{visualgenome}, adopted traces still concentrate attention inside the queried object at $1.5\times$ the density of the rest of the image for InternVL3.5-8B and $3.9\times$ for MiMo-VL-7B, against $3.1\times$ and $3.8\times$ for recovered traces (\cref{fig:rq3_modes_attn}c): stronger object attention accompanies recovery in InternVL3.5-8B, yet attending to the object does not by itself prevent adoption.
Adoption is thus not explained by a loss of attention: models keep attending to the queried object, while the thinking block rarely questions the written claim (probes~\cite{alain2017probes} and logit lens~\cite{logitlens} in \cref{sec:app_rq3}).

\subsection{Mitigating False-Thought Adoption}
\label{sec:rq3_intervene}

All results below use Set~A capable items (full breakdowns in \cref{sec:app_interventions}).

\paragraph{Decoding-time attention steering is ineffective.}
Several training-free methods mitigate multimodal hallucinations by penalizing over-trust in text tokens or amplifying attention to image tokens during decoding~\cite{opera,pai2024,kang2025sink,visualswap,laser}.
Consistent with our observation that adopting models already attend to the target object, simply boosting visual attention has limited effect.
Indeed, doubling image attention weights~\cite{visualswap} or applying the step-wise adaptive amplification of SAVAA~\cite{savaa} on InternVL3.5-8B and MiMo-VL-7B shifts recovery by at most $-2.7$ to $+4.7$ percentage points, leaving GQA adoption high at 36--52\% (\cref{tab:app_steering_full}).

\begin{table}[t]
\centering\footnotesize
\renewcommand{\arraystretch}{0.88}
\setlength{\tabcolsep}{4.0pt}
\resizebox{0.92\linewidth}{!}{%
\begin{tabular}{lccc}
\toprule
Model & GQA & ChartQA-H & MATH \\
\midrule
InternVL3.5-8B & $\mathbf{+15.1}$ / 98.8 & $+25.4$ / 89.6 & $+2.7$ / 86.2  \\
Qwen3.5-35B    & $+12.3$ / 97.4 & $+7.1$ / 92.9  & $-3.4$ / 96.6  \\
Qwen3.5-9B     & $+10.7$ / 98.0 & $+14.0$ / 87.9 & $-9.2$ / 95.0  \\
ERNIE-4.5-VL   & $+8.6$ / 90.6  & $\mathbf{+29.5}$ / 88.1 & $\mathbf{+13.5}$ / 85.1  \\
Qwen3-VL-30B   & $+8.6$ / 98.9  & $+4.1$ / 95.4           & $+3.5$ / 95.7  \\
Kimi-VL-A3B    & $+8.4$ / 95.5  & $+11.6$ / 77.6 & $+12.3$ / 85.5  \\
Gemma-4-26B    & $+6.2$ / 100.0 & $-0.9$ / 100.0          & $\phantom{+}0.0$ / 99.6  \\
Qwen3.6-35B    & $+4.6$ / 99.0           & $-14.5$ / 79.6 & $+5.4$ / 97.9  \\
GLM-4.1V-9B    & $+2.7$ / 99.5           & $-3.6$ / 86.2           & $-13.5$ / 84.1  \\
MiMo-VL-7B     & $+2.0$ / 97.5           & $-1.4$ / 95.2           & $+9.7$ / 90.3  \\
\bottomrule
\end{tabular}}
\caption{Check-cue prompting under a bare false statement across all 10 models on Set~A capable items $\mathcal{C}_A$ ($n=139\text{--}243$; Set~A+B in \cref{tab:app_c3_full}): $\Delta\mathrm{Rec}$ (pp) on $s^-$ / Keep-$a^+$ (\%) on $s^+$ ($100.0\%$ base). Bold: largest $\Delta\mathrm{Rec}$ per benchmark.}
\label{tab:checkcue}
\vspace{-7pt}
\end{table}

\paragraph{Check-cue prompting helps vulnerable models but induces over-correction.}
A common strategy to elicit self-correction or visual re-examination is to insert an explicit verification prompt or rethinking trigger~\cite{huang2024selfcorrect,kamoi2024selfcorrect,vlrethinker,reground,visualswap}.
Since adopting models still attend to the queried object, we test whether appending ``Let me check this condition against the image.'' right after the prefix moves the check into the thinking block.
The cue raises GQA recovery for all ten models, and ChartQA-H recovery by up to 30 points for low-recovery models such as ERNIE-4.5-VL and InternVL3.5-8B (\cref{tab:checkcue}).
However, generic verification cues introduce a trade-off~\cite{huang2024selfcorrect,kamoi2024selfcorrect,sherlock}: on multi-step tasks, the cue causes several stronger models, such as Qwen3.6-35B on ChartQA-H and GLM-4.1V-9B on MATH, to second-guess valid intermediate steps and even lowers accuracy when the injected statement $s^+$ is true.

\begin{table}[t]
\centering\footnotesize
\setlength{\tabcolsep}{2.3pt}
\begin{tabular}{lccccccc}
\toprule
 & \multicolumn{5}{c}{\Bare (trained framing)} & \multicolumn{2}{c}{\Look (unseen)} \\
\cmidrule(lr){2-6}\cmidrule(lr){7-8}
Arm & GQA & Chart & MATH & Rec. & Adopt & Rec. & Adopt \\
\midrule
\multicolumn{8}{l}{\textit{InternVL3.5-8B} ($|\mathcal{C}_A|=527$)} \\
Base          & 48.2 & 59.5 & 73.9 & 61.1 & 27.3 & 43.8 & 40.2 \\
Control ($C$)  & 57.8 & 71.1 & 78.7 & 69.6 & 12.9 & 63.9 & 22.0 \\
Recovery ($R$) & \textbf{80.7} & \textbf{90.2} & \textbf{93.6} & \textbf{88.4} & \textbf{4.7} & \textbf{87.3} & \textbf{7.6} \\
$R - C$ & $+22.9$ & $+19.1$ & $+14.9$ & $+18.8$ & $-8.2$ & $+23.3$ & $-14.4$ \\
\midrule
\multicolumn{8}{l}{\textit{MiMo-VL-7B} ($|\mathcal{C}_A|=600$)} \\
Base          & 47.5 & 91.3 & 71.3 & 70.3 & 21.5 & 54.8 & 35.8 \\
Control ($C$)  & 49.0 & 64.3 & 63.1 & 58.8 & 12.5 & 57.2 & 19.3 \\
Recovery ($R$) & \textbf{73.2} & \textbf{94.2} & \textbf{87.7} & \textbf{85.2} & \textbf{2.5} & \textbf{82.8} & \textbf{4.0} \\
$R - C$ & $+24.2$ & $+30.0$ & $+24.6$ & $+26.3$ & $-10.0$ & $+25.7$ & $-15.3$ \\
\bottomrule
\end{tabular}
\caption{LoRA fine-tuning on Set~B with \Bare recovery demonstrations ($R$) or matched true-prefix traces ($C$), evaluated on held-out Set~A capable items (\%). $R$ transfers to the unseen \Look framing. Full results in \cref{tab:app_sft_full}.}
\label{tab:sft}
\vspace{-7pt}
\end{table}

\paragraph{Fine-tuning on recovery demonstrations reliably teaches visual verification.}
Standard multimodal post-training predominantly exposes models to internally consistent reasoning traces~\cite{deepseekr1,mmcot,deepeyes}, and recent studies show that self-correction behaviors rarely emerge without targeted supervision~\cite{r3v,sherlock,octopus}.
To test whether models can learn visual verification from recovery demonstrations~\cite{sherlock,v1point,reground,octopus}, we fine-tune InternVL3.5-8B and MiMo-VL-7B with LoRA~\cite{lora} on Set~B using two matched arms (\cref{tab:sft}): a \emph{Recovery ($R$)} arm trained on traces that check $s^-$ against the image and recover $a^+$, and a \emph{Control ($C$)} arm trained on true-prefix ($s^+$) traces on the exact same items, with the prefix masked from the loss in both arms.
On held-out Set~A, recovery fine-tuning ($R$) substantially outperforms the control ($C$) across all three domains, raising overall bare-statement recovery from 61.1\% to 88.4\% on InternVL3.5-8B and from 70.3\% to 85.2\% on MiMo-VL-7B while cutting adoption below 5\%.
Crucially, although arm $R$ is trained only on bare statements, this verification transfers directly to the unseen first-person framing (\Look), boosting recovery to 87.3\% and 82.8\% while keeping true-prefix accuracy above 93\% (\cref{tab:app_sft_full}).

\section{Related Work}
\label{sec:related}

\paragraph{Multimodal chain-of-thought reasoning and self-correction.}
Chain-of-thought reasoning~\cite{cot,deepseekr1} has led to reasoning VLMs that write an explicit thinking trace before answering~\cite{mmcot,vlrethinker,deepeyes}.
Whether these traces can be trusted is less clear.
In text-only LLMs, intrinsic self-correction is unreliable~\cite{huang2024selfcorrect,kamoi2024selfcorrect} and written rationales can be unfaithful to the computation behind the answer~\cite{lanham2023measuring,turpin2023unfaithful}, although reasoning models often flag errors injected into their chain of thought~\cite{vonrecum}.
Multimodal traces frequently contain noisy or hallucinated steps~\cite{r3v,morethinkinglessseeing,morethought,kancheti2026cot}, and revising them takes dedicated reflection or rollout training~\cite{r3v,vlrethinker,sherlock,octopus,tang2026vrrl}.
Related failures have been reported elsewhere: reasoning errors injected into video models are corrected in under 10\% of cases~\cite{zhu2026textual}, early predictions are reinforced rather than revised~\cite{sanchezvillegas2026answer}, stale chain-of-thought evidence competes with fresh visual recomputation~\cite{fan2026recompute}, incorrect reasoning traces degrade open-source multimodal reasoning models~\cite{zhao2026robustness}, and LVLMs accept their own earlier hallucinations despite the image~\cite{zhong2024snowball,jiang2026mmsnowball}.
These studies mostly let errors arise freely.
To isolate the check itself, our work pre-fills a single false claim at the start of the thinking block and asks whether the model catches it, using the image or using equivalent text.

\paragraph{Cross-modal conflicts and modality preference.}
Language models are easily swayed by distracting or conflicting context~\cite{shi2023distracted,xie2024chameleon} and by user sycophancy~\cite{sharma2024sycophancy}.
VLMs show similar weaknesses.
POPE~\cite{pope} and HallusionBench~\cite{hallusionbench} find that they hallucinate under language priors or visual illusions~\cite{vcd}, and they tend to favor conflicting text in the prompt~\cite{wordsorvision} or semantic anchors~\cite{vlmsneedwords} over the image.
Isomorphic benchmarks such as IsoBench~\cite{isobench}, MathVerse~\cite{mathverse}, \mbox{SpatialEval}~\cite{spatialeval}, and Prism~\cite{prism} further show that models reason better over text than over images~\cite{simpletohard,composeandfuse}.
In these settings, however, the conflicting text comes from the user or the input rather than from the model's own reasoning.
To this end, our work places the false claim inside the thinking and pairs image and text evidence under the same claim, so that deferring to one's own thinking can be compared with complying with the user, and checking against an image with checking against text.

\paragraph{Visual grounding and attention dynamics in long generation.}
Mechanistic studies have examined how visual information flows across layers~\cite{neo2025interpreting,crossmodalflow,ashok2025recall,toolate} and why grounding weakens during generation.
Hallucinations have been attributed to over-trust in summary tokens~\cite{opera}, attention sinks~\cite{kang2025sink}, associative steps~\cite{ma2026mcotdivergent}, and decaying visual attention over long reasoning traces~\cite{tvc,morethinkinglessseeing,morethought,laser,jin2026looklight}.
These findings have motivated attention amplification and over-trust penalties during decoding~\cite{opera,pai2024,kang2025sink,laser}, feature re-injection~\cite{tvc}, token pointing~\cite{v1point}, and self-diagnosed re-examination~\cite{reground}.
VisualSwap~\cite{visualswap} swaps the image partway through reasoning and finds that models verbalize re-reading cues without updating their beliefs.
We keep the image unchanged and instead plant a false claim in the model's own thinking, and test whether lost visual attention explains the failure to correct it.

\section{Conclusion}
\label{sec:discussion}

We investigated whether reasoning VLMs use visual evidence to verify and correct false visual statements in their own thinking traces.
Across a 3{,}000-item benchmark and ten open-weight models, we found that models frequently adopt false written thoughts over clear visual evidence, even on items solved with near certainty without intervention, while recovering far more often from text.
Recovery is modulated by task domain, statement framing, prompt placement, and model scale; models adopting the false claim still attend to the queried object but rarely question it while thinking.
Consistent with this diagnosis, attention steering is ineffective and check-cue prompting induces over-correction, whereas recovery fine-tuning improves visual verification and transfers to an unseen framing, pointing to conflict-resolution post-training for robust multimodal reasoning.

\section*{LLM Usage Statement}
The authors conceived the research, designed the experiments, and verified all analyses, citations, and claims. LLMs were used in three ways. (1)~Data construction (\cref{sec:setup}): GPT rendered the MATH problems as page images, while GPT and Gemini drafted and cross-checked candidate statement pairs and text evidence, which the authors audited. (2)~Evaluation: GPT and Claude adjudicated answers that rule-based matching could not resolve, and Gemini and Claude independently labeled reasoning behavior in the 3{,}000-trace sample of \cref{sec:rq3_mech}. (3)~Code and writing: Claude and Gemini assisted with coding, plotting, and drafting and editing prose. The authors take full responsibility for the content.

\clearpage
\onecolumn
\appendix
\raggedbottom
\makeatletter
\setlength{\@fptop}{0pt}
\setlength{\@fpsep}{10pt plus 2pt}
\setlength{\@fpbot}{0pt plus 1fil}
\setlength{\@dblfptop}{0pt}
\setlength{\@dblfpsep}{10pt plus 2pt}
\setlength{\@dblfpbot}{0pt plus 1fil}
\makeatother
\renewcommand{\topfraction}{0.95}
\renewcommand{\bottomfraction}{0.95}
\renewcommand{\dbltopfraction}{0.95}
\renewcommand{\textfraction}{0.05}
\renewcommand{\floatpagefraction}{0.60}
\renewcommand{\dblfloatpagefraction}{0.60}
\setlength{\dbltextfloatsep}{10pt plus 2pt minus 2pt}
\setlength{\textfloatsep}{10pt plus 2pt minus 2pt}
\setcounter{topnumber}{4}
\setcounter{bottomnumber}{2}
\setcounter{totalnumber}{6}
\setcounter{dbltopnumber}{4}
\begin{center}
{\Large\bfseries Appendix}
\end{center}
\vspace{0.5em}

\FloatBarrier
\section{Benchmark Construction, Subtypes, and Scoring Audit}
\label{sec:app_details}

\begin{table*}[!htbp]
\centering\footnotesize
\setlength{\tabcolsep}{4.5pt}
\renewcommand{\arraystretch}{1.18}
\begin{tabular}{|>{\raggedright\arraybackslash}p{3.0cm}|m{4.9cm}|>{\raggedright\arraybackslash}m{4.5cm}|>{\raggedright\arraybackslash}m{3.8cm}|}
\hline
\multicolumn{2}{|m{8.1cm}|}{\centering\vspace{3pt}\includegraphics[height=3.6cm,keepaspectratio]{fig/app_ex_gqa.jpg}\vspace{3pt}} &
\textbf{Domain \& ID:} GQA (\texttt{011027149})\newline
\textbf{Q:} The shorts have what color? &
\textbf{GT ($a^+$):} black\newline
\textbf{False ($a^-$):} blue \\
\hline
\textbf{True Prefix ($s^+$):} & \multicolumn{3}{p{13.6cm}|}{\texttt{<think>}The tennis player's shorts are black.} \\
\hline
\textbf{False Prefix ($s^-$):} & \multicolumn{3}{p{13.6cm}|}{\texttt{<think>}The tennis player's shorts are blue.} \\
\hline
\textbf{Paired Text ($T$):} & \multicolumn{3}{p{13.6cm}|}{A male tennis player leans forward in the follow-through of a swing, gripping a racket in front of his body. He wears a loose white T-shirt with green lettering and \textbf{black shorts}, with a small section of his abdomen exposed as the shirt lifts. [\dots] The racket has a blue grip, a dark frame with white markings, and yellow strings. A dark blue windscreen or court wall, blue netting, a horizontal court net, and a narrow green playing surface form the background.} \\
\hline\hline
\multicolumn{2}{|m{8.1cm}|}{\centering\vspace{3pt}\includegraphics[width=7.8cm,height=4.0cm,keepaspectratio]{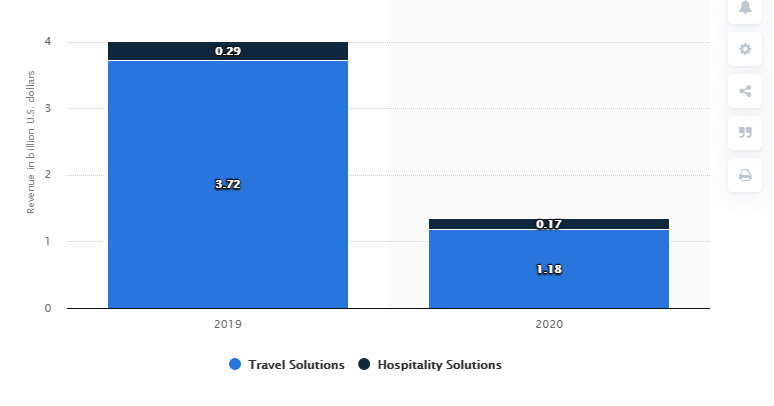}\vspace{3pt}} &
\textbf{Domain \& ID:} ChartQA-H (\texttt{multi\_col\_40183})\newline
\textbf{Q:} What is sum of the dark blue bar from 2019 to 2020? &
\textbf{GT ($a^+$):} $0.46$\newline
\textbf{False ($a^-$):} $0.57$ \\
\hline
\textbf{True Prefix ($s^+$):} & \multicolumn{3}{p{13.6cm}|}{\texttt{<think>}The Hospitality Solutions revenue in 2019 is 0.29 billion U.S. dollars.} \\
\hline
\textbf{False Prefix ($s^-$):} & \multicolumn{3}{p{13.6cm}|}{\texttt{<think>}The Hospitality Solutions revenue in 2019 is 0.40 billion U.S. dollars.} \\
\hline
\textbf{Paired Text ($T$):} & \multicolumn{3}{p{13.6cm}|}{Revenue by business segment, 2019 and 2020. Axis: Revenue in billion U.S. dollars. Blue = Travel Solutions; dark blue = Hospitality Solutions. \quad \texttt{| Year | Travel | Hospitality | 2019 | 3.72 | \textbf{0.29} | 2020 | 1.18 | \textbf{0.17} |}} \\
\hline\hline
\multicolumn{2}{|m{8.1cm}|}{\centering\vspace{4pt}\includegraphics[width=7.8cm,keepaspectratio]{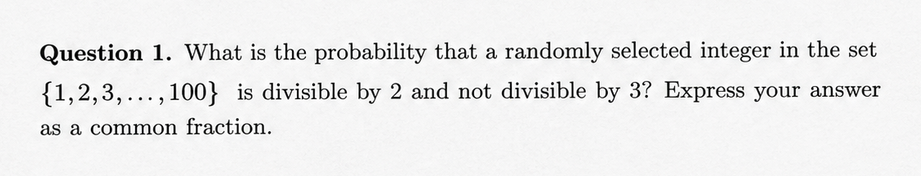}\vspace{4pt}} &
\textbf{Domain \& ID:} Rendered MATH (\texttt{prealg\_1655})\newline
\textbf{Q:} Solve the problem shown in the image. &
\textbf{GT ($a^+$):} $17/50$\newline
\textbf{False ($a^-$):} $42/125$ \\
\hline
\textbf{True Prefix ($s^+$):} & \multicolumn{3}{p{13.6cm}|}{\texttt{<think>}The largest integer in the uniform sampling set is 100.} \\
\hline
\textbf{False Prefix ($s^-$):} & \multicolumn{3}{p{13.6cm}|}{\texttt{<think>}The largest integer in the uniform sampling set is 125.} \\
\hline
\textbf{Paired Text ($T$):} & \multicolumn{3}{p{13.6cm}|}{What is the probability that a randomly selected integer in the set $\mathbf{\{1, 2, 3, \ldots, 100\}}$ is divisible by 2 and not divisible by 3? Express your answer as a common fraction.} \\
\hline
\end{tabular}
\caption{Illustration of benchmark items across GQA, ChartQA-H, and rendered MATH showing the input image $I$, question $q$, ground-truth and counterfactual answers $(a^+, a^-)$, injected thinking prefixes $(s^+, s^-)$ under \Bare, and paired textual evidence $T$ used in the text-only modality comparison (\cref{sec:app_rq2}). On GQA, the question-blind scene description $T$ embeds the true target attribute alongside distractor objects sharing the counterfactual color (\eg, blue grip, blue windscreen, and blue netting).}
\label{tab:app_examples}
\end{table*}

\begin{table*}[!htbp]
\centering\footnotesize
\renewcommand{\arraystretch}{0.92}
{\includegraphics[width=0.90\textwidth]{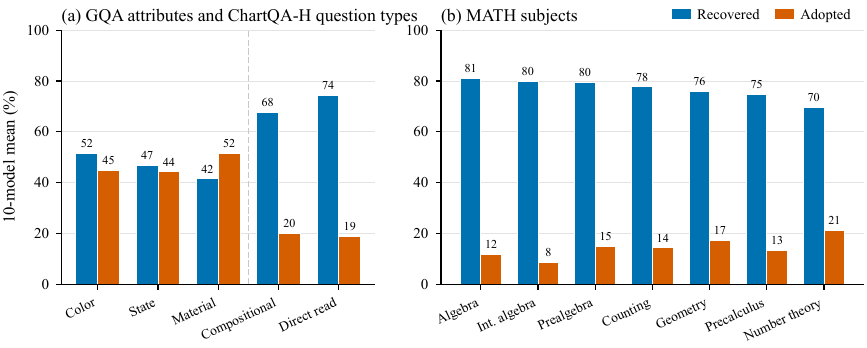}
\vspace{-3pt}
\captionof{figure}{Fine-grained conditional recovery ($a^+$, blue) and adoption ($a^-$, orange) under \Bare averaged across all 10 models across the 3{,}000-item benchmark (1{,}000 per domain). (a) GQA attribute subtypes (Color, State/Other, Material) and ChartQA-H reasoning types (Compositional vs.\ Direct Readout). (b) Rendered MATH across all seven competition subjects sorted by mean recovery rate.}
\label{fig:app_subtypes}}
\end{table*}

\begin{table*}[!htbp]
\centering\footnotesize
\renewcommand{\arraystretch}{0.92}
\setlength{\tabcolsep}{6.0pt}
\begin{tabular}{llcccc}
\toprule
Domain & Subtype & Items ($N$) & 10-Model Mean $|\mathcal{C}|$ & Conditional Rec.\ Rate [Min, Max] (\%) & Conditional Adopt.\ Rate (\%) \\
\midrule
\multirow{3}{*}{GQA}
 & Color attribute       & 706 & 583.1 & 51.7 [17.1, 86.1] & 44.8 \\
 & State / condition     & 157 &  95.6 & 46.8 [14.4, 72.9] & 44.3 \\
 & Material attribute    & 137 & 109.7 & 41.5 [0.0, 82.9]  & 51.6 \\
\midrule
\multirow{2}{*}{ChartQA-H}
 & Compositional         & 786 & 608.9 & 67.7 [36.7, 98.3] & 20.1 \\
 & Direct readout        & 214 & 168.8 & 74.4 [44.4, 99.0] & 18.8 \\
\midrule
\multirow{7}{*}{MATH}
 & Algebra               & 203 & 189.5 & 81.0 [53.7, 100.0] & 11.9 \\
 & Interm.\ Algebra      & 151 & 121.9 & 80.0 [60.8, 100.0] &  8.5 \\
 & Prealgebra            & 158 & 150.4 & 79.6 [48.3, 100.0] & 15.0 \\
 & Counting \& Prob.     & 121 & 106.2 & 77.7 [45.7, 100.0] & 14.4 \\
 & Geometry              & 110 &  96.6 & 76.1 [48.9, 99.1]  & 17.4 \\
 & Precalculus           & 119 &  95.2 & 74.7 [50.0, 99.1]  & 13.4 \\
 & Number Theory         & 138 & 119.2 & 69.7 [40.9, 99.3]  & 21.4 \\
\bottomrule
\end{tabular}
\caption{Fine-grained subtype breakdown across the 3{,}000 benchmark items (1{,}000 per domain) under \Bare. We report total items $N$, 10-model mean symmetric capable-set size $|\mathcal{C}|$ matching \cref{tab:app_w2_gqa,tab:app_w2_chart,tab:app_w2_math}, 10-model mean conditional recovery rate (\%) with min--max model range, and 10-model mean conditional adoption rate (\%).}
\label{tab:app_subtypes}
\end{table*}

\begin{table*}[!htbp]
\centering\footnotesize
\renewcommand{\arraystretch}{0.92}
\setlength{\tabcolsep}{7.0pt}
\begin{tabular}{llcl}
\toprule
Model & Architecture & Reasoning Opener & Hugging Face Checkpoint \\
\midrule
Gemma-4-26B    & MoE (26B total / 4B active)  & \texttt{<|channel>thought\textbackslash n} & \nolinkurl{google/gemma-4-26b-a4b-it} \\
Qwen3-VL-30B   & MoE (30B total / 3B active)  & \texttt{<think>}\texttt{\textbackslash n}   & \nolinkurl{Qwen/Qwen3-VL-30B-A3B-Thinking} \\
Qwen3.6-35B    & MoE (35B total / 3B active)  & \texttt{<think>}\texttt{\textbackslash n}   & \nolinkurl{Qwen/Qwen3.6-35B-A3B} \\
Qwen3.5-35B    & MoE (35B total / 3B active)  & \texttt{<think>}\texttt{\textbackslash n}   & \nolinkurl{Qwen/Qwen3.5-35B-A3B} \\
GLM-4.1V-9B    & Dense (9B)                   & \texttt{<think>}\texttt{\textbackslash n}   & \nolinkurl{THUDM/GLM-4.1V-9B-Thinking} \\
Qwen3.5-9B     & Dense (9B)                   & \texttt{<think>}\texttt{\textbackslash n}   & \nolinkurl{Qwen/Qwen3.5-9B} \\
InternVL3.5-8B & Dense (8B)                   & \texttt{<think>}\texttt{\textbackslash n}   & \nolinkurl{OpenGVLab/InternVL3_5-8B} \\
MiMo-VL-7B     & Dense (7B)                   & \texttt{<think>}\texttt{\textbackslash n}   & \nolinkurl{XiaomiMiMo/MiMo-VL-7B-RL} \\
Kimi-VL-A3B    & MoE (16B total / 3B active)  & \texttt{$\triangleleft$think$\triangleright$} & \nolinkurl{moonshotai/Kimi-VL-A3B-Thinking-2506} \\
ERNIE-4.5-VL   & MoE (28B total / 3B active)  & \texttt{<think>}\texttt{\textbackslash n}   & \nolinkurl{PaddlePaddle/ERNIE-4.5-VL-28B-A3B-Thinking} \\
\midrule
\multicolumn{4}{l}{\textit{Additional Qwen3.5 checkpoints evaluated in the scaling analysis (\cref{sec:app_wording}):}} \\
Qwen3.5-4B     & Dense (4B)                   & \texttt{<think>}\texttt{\textbackslash n}   & \nolinkurl{Qwen/Qwen3.5-4B} \\
Qwen3.5-2B     & Dense (2B)                   & \texttt{<think>}\texttt{\textbackslash n}   & \nolinkurl{Qwen/Qwen3.5-2B} \\
\bottomrule
\end{tabular}
\caption{Summary of evaluated open-weight reasoning VLMs, including model architecture, total and active parameter counts, native reasoning delimiters, and official Hugging Face checkpoints.}
\label{tab:app_models}
\end{table*}

\paragraph{Benchmark construction.}
Each of the 3{,}000 items pairs a true statement and answer $(s^+, a^+)$ with a counterfactual $(s^-, a^-)$, $a^- \neq a^+$ (\cref{tab:app_examples}).
On GQA~\cite{gqa}, $s^-$ replaces the queried attribute with another from the same category; on ChartQA-H~\cite{chartqa}, it changes one plotted value so that the required operation yields $a^-$; on rendered MATH~\cite{mathdataset}, the problem appears only as a page image, and $s^-$ misstates one printed quantity.
The first 1{,}500 items (500 per domain) are split into two stratified halves of 750 items: held-out evaluation Set~A and training Set~B.
\Cref{tab:app_subtypes} and \cref{fig:app_subtypes} break the results down by subtype.

\paragraph{Prefix wordings.}
For a target fact $f$ (\eg, \emph{``the walls of the bathroom are blue''}), the prefixes are:
{\raggedright
\begin{itemize}[leftmargin=*, itemsep=2pt, topsep=3pt]
\item \textbf{\Bare (Primary, $s^+/s^-$):} \texttt{<think>}The walls of the bathroom are blue.\texttt{\textbackslash n}
\item \textbf{\Neutral ($s^{\text{neut}}$):} \texttt{<think>}Let me carefully examine the image to answer the question.\texttt{\textbackslash n}
\item \textbf{\Look:} \texttt{<think>}Looking at the image, the walls of the bathroom are blue.\texttt{\textbackslash n}
\item \textbf{\Cond:} \texttt{<think>}The problem gives the following condition: the walls of the bathroom are blue.\texttt{\textbackslash n}
\item \textbf{\Bare + Check-Cue:} \texttt{<think>}The walls of the bathroom are blue. Let me check this condition against the image.\texttt{\textbackslash n}
\end{itemize}
}

\paragraph{Models and scoring.}
\Cref{tab:app_models} lists the checkpoints and their reasoning openers.
Each model generates one sample per item with its official sampling configuration and a 32{,}768-token budget in vLLM~\cite{vllm}.
Answers are matched to $(a^+, a^-)$ by alias and numeric normalization, a rule-based answer extractor, and a blind adjudication pass that sees neither the condition nor the thinking trace; unresolved or truncated responses count as \emph{other}.

\FloatBarrier
\section{Complete \Bare Visual Recovery Results by Domain}
\label{sec:app_w2}
\paragraph{Per-domain results.}
\Cref{tab:app_w2_gqa,tab:app_w2_chart,tab:app_w2_math} report all four conditions per domain: unconditional accuracy over all 1{,}000 items on the left and capable-set metrics on $\mathcal{C}$ on the right, with 95\% item-level bootstrap intervals from 2{,}000 resamples.

\paragraph{Confidence and consensus on GQA.}
\Cref{tab:app_gqa_confidence} restricts GQA to items that each model perceives confidently or that all ten models solve.
Confidence is the No-Prefix final-answer token probability $P(a^+\mid I,q,\emptyset)$, recorded on the 500 GQA items of Set~A+B; the consensus set contains the 544 items that all ten models answer correctly without a prefix.
On these subsets, recovery and adoption stay close to their values on the full capable set.

\begin{table*}[!htbp]
\centering\footnotesize
\setlength{\tabcolsep}{1.5pt}
\begin{tabular}{lcccccc@{\hspace{8pt}}cccc}
\toprule
 & \multicolumn{6}{c}{Unconditional Metrics (All 1{,}000 GQA Items)} & \multicolumn{4}{c}{Conditional Metrics on Capable Set $\mathcal{C}$} \\
\cmidrule(lr){2-7}\cmidrule(lr){8-11}
Model & No-Pref $a^+$ & Neut.\ $a^+$ & $s^+$ $a^+$ & $s^-$ $a^+$ & $s^-$ $a^-$ & Drop (pp) [95\% CI] & $|\mathcal{C}|$ & Neut.\ Keep & Rec.\ Rate [95\% CI] & Adopt.\ Rate [95\% CI] \\
\midrule
Gemma-4-26B     & 84.6 & 87.8 & 96.5 & 74.3 & 16.0 & 22.2 [19.6, 24.9] & 843 & 96.2 & 84.0 [81.5, 86.4] & 12.8 [10.6, 15.1] \\
Qwen3-VL-30B    & 81.9 & 83.3 & 98.2 & 56.9 & 34.6 & 41.3 [38.1, 44.4] & 816 & 92.0 & 64.7 [61.4, 67.9] & 32.6 [29.4, 35.9] \\
Qwen3.6-35B     & 84.7 & 89.3 & 96.9 & 53.1 & 37.7 & 43.8 [40.7, 46.8] & 836 & 96.5 & 59.0 [55.6, 62.3] & 35.6 [32.4, 38.9] \\
GLM-4.1V-9B     & 82.8 & 83.5 & 96.7 & 50.4 & 41.0 & 46.3 [43.2, 49.3] & 823 & 93.8 & 57.8 [54.2, 61.1] & 38.4 [35.2, 41.9] \\
MiMo-VL-7B      & 81.9 & 83.6 & 96.7 & 50.1 & 39.0 & 46.6 [43.4, 49.7] & 816 & 92.9 & 57.2 [53.8, 60.4] & 35.7 [32.6, 39.1] \\
Qwen3.5-35B     & 82.9 & 86.9 & 96.9 & 47.7 & 45.3 & 49.2 [46.1, 52.3] & 824 & 94.8 & 54.0 [50.6, 57.4] & 42.5 [39.1, 46.0] \\
InternVL3.5-8B  & 76.4 & 80.3 & 96.4 & 37.7 & 54.1 & 58.7 [55.6, 61.7] & 756 & 93.5 & 46.0 [42.5, 49.6] & 50.9 [47.1, 54.5] \\
Qwen3.5-9B      & 83.1 & 87.2 & 97.5 & 35.4 & 58.1 & 62.1 [59.0, 65.1] & 829 & 94.5 & 39.0 [35.6, 42.3] & 57.3 [53.7, 60.6] \\
Kimi-VL-A3B     & 79.3 & 83.9 & 97.6 & 16.5 & 80.2 & 81.1 [78.6, 83.6] & 786 & 94.3 & 19.8 [17.0, 22.6] & 78.8 [75.8, 81.6] \\
ERNIE-4.5-VL    & 79.4 & 81.9 & 70.6 & 12.0 & 62.4 & 58.6 [55.3, 62.0] & 555 & 94.2 & 16.2 [13.3, 19.5] & 75.0 [71.4, 78.6] \\
\midrule
\textbf{10-Model Mean} & \textbf{81.7} & \textbf{84.8} & \textbf{94.4} & \textbf{43.4} & \textbf{46.8} & \textbf{51.0} & \textbf{788.4} & \textbf{94.3} & \textbf{49.8} & \textbf{46.0}  \\
\bottomrule
\end{tabular}
\caption{Complete visual recovery results on GQA (1{,}000 natural-scene items) under \Bare across all four conditions ($\emptyset, s^{\text{neut}}, s^+, s^-$). Here $\mathcal{C} = \{x : \hat{a}^{(\emptyset)}(x) = a^+ \wedge \hat{a}^{(s^+)}(x) = a^+\}$ is the symmetric capable set. Each unconditional column gives the share (\%) of items answered $a^+$ under the stated prefix, except $s^-$ $a^-$, the share answered $a^-$; Drop is the unconditional accuracy drop (pp) from $s^+$ to $s^-$, and Neut.\ Keep is accuracy retention (\%) under $s^{\text{neut}}$ on $\mathcal{C}$.}
\label{tab:app_w2_gqa}
\label{tab:app_w2}
\end{table*}

\begin{table*}[!htbp]
\centering\footnotesize
\setlength{\tabcolsep}{1.5pt}
\begin{tabular}{lcccccc@{\hspace{8pt}}cccc}
\toprule
 & \multicolumn{6}{c}{Unconditional Metrics (All 1{,}000 ChartQA-H Items)} & \multicolumn{4}{c}{Conditional Metrics on Capable Set $\mathcal{C}$} \\
\cmidrule(lr){2-7}\cmidrule(lr){8-11}
Model & No-Pref $a^+$ & Neut.\ $a^+$ & $s^+$ $a^+$ & $s^-$ $a^+$ & $s^-$ $a^-$ & Drop (pp) [95\% CI] & $|\mathcal{C}|$ & Neut.\ Keep & Rec.\ Rate [95\% CI] & Adopt.\ Rate [95\% CI] \\
\midrule
Gemma-4-26B     & 92.9 & 93.1 & 93.1 & 92.4 &  0.9 &  0.7 [$-$0.4, $+$1.8] & 919 & 99.5 & 98.5 [97.6, 99.2] &  1.0 [0.4, 1.6] \\
MiMo-VL-7B      & 88.7 & 85.9 & 85.5 & 77.2 &  8.7 &  8.3 [6.0, 10.7]      & 823 & 96.4 & 87.7 [85.5, 89.9] &  8.3 [6.4, 10.2] \\
Qwen3-VL-30B    & 86.5 & 85.2 & 83.7 & 71.4 & 11.4 & 12.3 [9.4, 15.1]      & 795 & 94.7 & 80.8 [78.1, 83.4] & 11.9 [9.8, 14.2] \\
Qwen3.6-35B     & 92.3 & 91.4 & 89.1 & 72.6 & 11.9 & 16.5 [13.7, 19.2]     & 871 & 98.9 & 78.4 [75.7, 81.1] & 12.7 [10.7, 15.0] \\
GLM-4.1V-9B     & 80.0 & 71.4 & 75.0 & 62.9 &  9.5 & 12.1 [9.2, 15.0]      & 690 & 88.8 & 78.4 [75.4, 81.6] & 10.1 [8.0, 12.5] \\
Qwen3.5-35B     & 86.5 & 92.4 & 88.9 & 67.7 & 13.4 & 21.2 [18.3, 24.2] & 826 & 98.5 & 73.8 [70.8, 76.9] & 13.6 [11.3, 16.0] \\
Qwen3.5-9B      & 88.9 & 91.1 & 87.3 & 53.2 & 23.8 & 34.1 [30.8, 37.4] & 826 & 98.3 & 57.4 [54.1, 60.7] & 25.9 [23.0, 28.8] \\
InternVL3.5-8B  & 84.7 & 82.0 & 73.5 & 46.1 & 23.4 & 27.4 [24.1, 30.7] & 685 & 94.2 & 56.6 [53.0, 60.4] & 27.2 [23.6, 30.4] \\
Kimi-VL-A3B     & 79.4 & 80.2 & 67.8 & 33.7 & 29.4 & 34.1 [30.4, 38.0] & 619 & 94.8 & 40.9 [36.8, 44.9] & 37.8 [34.1, 41.8] \\
ERNIE-4.5-VL    & 88.4 & 87.7 & 75.7 & 33.8 & 40.7 & 41.9 [38.4, 45.4] & 723 & 95.6 & 38.7 [35.1, 42.5] & 49.8 [46.1, 53.7] \\
\midrule
\textbf{10-Model Mean} & \textbf{86.8} & \textbf{86.0} & \textbf{82.0} & \textbf{61.1} & \textbf{17.3} & \textbf{20.9} & \textbf{777.7} & \textbf{96.0} & \textbf{69.1} & \textbf{19.8}  \\
\bottomrule
\end{tabular}
\caption{Complete visual recovery results on ChartQA-H (1{,}000 human-authored chart reasoning items) under \Bare across all four conditions ($\emptyset, s^{\text{neut}}, s^+, s^-$), sorted by conditional recovery rate on the symmetric capable set $\mathcal{C}$.}
\label{tab:app_w2_chart}
\end{table*}

\begin{table*}[!htbp]
\centering\footnotesize
\setlength{\tabcolsep}{1.5pt}
\begin{tabular}{lcccccc@{\hspace{8pt}}cccc}
\toprule
 & \multicolumn{6}{c}{Unconditional Metrics (All 1{,}000 Rendered MATH Items)} & \multicolumn{4}{c}{Conditional Metrics on Capable Set $\mathcal{C}$} \\
\cmidrule(lr){2-7}\cmidrule(lr){8-11}
Model & No-Pref $a^+$ & Neut.\ $a^+$ & $s^+$ $a^+$ & $s^-$ $a^+$ & $s^-$ $a^-$ & Drop (pp) [95\% CI] & $|\mathcal{C}|$ & Neut.\ Keep & Rec.\ Rate [95\% CI] & Adopt.\ Rate [95\% CI] \\
\midrule
Gemma-4-26B     & 98.1 & 98.1 & 98.8 & 98.7 &  0.0 &  0.1 [$-$0.3, $+$0.6] & 979 & 99.4 & 99.7 [99.4, 100.0] &  0.0 [0.0, 0.0] \\
Qwen3.5-35B     & 96.9 & 98.3 & 98.7 & 91.9 &  4.3 &  6.8 [5.2, 8.5]       & 963 & 99.4 & 92.7 [91.1, 94.3]  &  4.4 [3.1, 5.6] \\
Qwen3-VL-30B    & 95.8 & 92.1 & 94.9 & 89.8 &  5.2 &  5.1 [3.2, 7.1]       & 927 & 95.0 & 92.3 [90.6, 94.1]  &  5.3 [3.9, 6.7] \\
Qwen3.5-9B      & 96.0 & 97.9 & 98.5 & 83.5 &  9.6 & 15.0 [12.7, 17.5]     & 953 & 99.0 & 84.5 [82.1, 86.8]  &  9.8 [7.9, 11.6] \\
Qwen3.6-35B     & 98.5 & 92.2 & 98.8 & 81.9 & 12.4 & 16.9 [14.5, 19.3] & 979 & 93.3 & 82.4 [80.0, 84.8] & 12.7 [10.6, 14.9] \\
GLM-4.1V-9B     & 90.2 & 69.6 & 87.3 & 70.2 & 16.5 & 17.1 [14.3, 19.9] & 845 & 78.9 & 75.7 [72.9, 78.7] & 17.2 [14.7, 19.6] \\
MiMo-VL-7B      & 94.0 & 76.5 & 79.1 & 58.9 & 18.8 & 20.2 [17.1, 23.4] & 770 & 90.3 & 69.0 [65.7, 72.2] & 21.6 [18.7, 24.4] \\
InternVL3.5-8B  & 96.4 & 91.0 & 78.2 & 57.9 & 12.1 & 20.3 [17.3, 23.6] & 771 & 96.2 & 67.7 [64.2, 71.1] & 13.5 [11.2, 16.0] \\
Kimi-VL-A3B     & 92.2 & 80.0 & 75.8 & 53.2 & 18.7 & 22.6 [19.2, 26.1] & 732 & 88.7 & 61.6 [58.2, 64.9] & 20.9 [17.9, 23.9] \\
ERNIE-4.5-VL    & 95.4 & 90.8 & 89.1 & 48.8 & 35.9 & 40.3 [37.0, 43.6] & 871 & 95.9 & 50.7 [47.4, 54.1] & 38.0 [34.8, 41.2] \\
\midrule
\textbf{10-Model Mean} & \textbf{95.3} & \textbf{88.6} & \textbf{89.9} & \textbf{73.5} & \textbf{13.3} & \textbf{16.4} & \textbf{879.0} & \textbf{93.6} & \textbf{77.6} & \textbf{14.3}  \\
\bottomrule
\end{tabular}
\caption{Complete visual recovery results on Rendered MATH (1{,}000 printed competition math problems) under \Bare across all four conditions ($\emptyset, s^{\text{neut}}, s^+, s^-$), sorted by conditional recovery rate on the symmetric capable set $\mathcal{C}$.}
\label{tab:app_w2_math}
\end{table*}

\begin{table*}[!htbp]
\centering\footnotesize
\setlength{\tabcolsep}{3.4pt}
\begin{tabular}{lccc@{\hspace{8pt}}ccc@{\hspace{8pt}}ccc@{\hspace{8pt}}cccc}
\toprule
 & \multicolumn{3}{c}{All Capable $\mathcal{C}$ (1{,}000 GQA)} & \multicolumn{3}{c}{High Conf.\ ($P(a^+\mid\emptyset)\ge 0.90$)} & \multicolumn{3}{c}{Near-Certain ($P(a^+\mid\emptyset)\ge 0.99$)} & \multicolumn{4}{c}{10/10 Consensus ($N=544$)} \\
\cmidrule(lr){2-4}\cmidrule(lr){5-7}\cmidrule(lr){8-10}\cmidrule(lr){11-14}
Model & $|\mathcal{C}|$ & Rec.\ & Adopt.\ & $n$ & Rec.\ & Adopt.\ & $n$ & Rec.\ & Adopt.\ & $n$ & Rec.\ & Adopt.\ & $\Delta_{\text{txt}-\text{img}}$ \\
\midrule
Gemma-4-26B    & 843 & 84.0 & 12.8 & 374 & 82.6 & 13.1 & 322 & 82.6 & 13.0 & 543 & 90.1 & 8.7 & $+8.9$ \\
Qwen3-VL-30B   & 816 & 64.7 & 32.6 & 293 & 65.5 & 31.4 & 131 & 66.4 & 30.5 & 544 & 73.5 & 25.0 & $+24.6$ \\
Qwen3.6-35B    & 836 & 59.0 & 35.6 & 306 & 55.6 & 39.2 & 159 & 54.7 & 39.0 & 543 & 67.6 & 29.3 & $-14.4$ \\
GLM-4.1V-9B    & 823 & 57.8 & 38.4 & 356 & 57.9 & 37.6 & 294 & 59.5 & 38.4 & 543 & 67.2 & 30.8 & $+26.2$ \\
MiMo-VL-7B     & 816 & 57.2 & 35.7 & 206 & 45.1 & 45.1 &  87 & 36.8 & 51.7 & 542 & 67.7 & 28.6 & $+30.3$ \\
Qwen3.5-35B    & 824 & 54.0 & 42.5 & 355 & 50.1 & 45.1 & 189 & 46.6 & 46.0 & 543 & 60.4 & 38.9 & $+24.5$ \\
InternVL3.5-8B & 756 & 46.0 & 50.9 & 290 & 41.0 & 56.9 & 212 & 43.9 & 54.7 & 542 & 51.5 & 45.6 & $+43.4$ \\
Qwen3.5-9B     & 829 & 39.0 & 57.3 & 221 & 43.9 & 50.7 &  43 & 53.5 & 41.9 & 544 & 44.5 & 53.5 & $+52.0$ \\
Kimi-VL-A3B    & 786 & 19.8 & 78.8 & 127 & 15.0 & 81.9 &  25 & 20.0 & 80.0 & 539 & 22.8 & 75.9 & $+66.9$ \\
ERNIE-4.5-VL   & 555 & 16.2 & 75.0 & 240 & 11.2 & 81.7 &  44 & 11.4 & 79.5 & 388 & 19.6 & 71.4 & $+43.2$ \\
\midrule
\textbf{10-Model Mean} & \textbf{788.4} & \textbf{49.8} & \textbf{46.0} & \textbf{276.8} & \textbf{46.8} & \textbf{48.3} & \textbf{150.6} & \textbf{47.5} & \textbf{47.5} & \textbf{527.1} & \textbf{56.5} & \textbf{40.8} & $\mathbf{+30.6}$  \\
\bottomrule
\end{tabular}
\caption{Perceptual confidence on GQA under \Bare: all capable items $\mathcal{C}$ over the 1{,}000 GQA questions; capable items whose No-Prefix answer probability $P(a^+\mid I,q,\emptyset)$ is at least 0.90 or 0.99, among the 500 GQA items of Set~A+B where token log-probabilities were recorded; and the 544 items that all ten models answer correctly without a prefix ($10/10$ consensus). In every block, $n$ counts the model's capable items in that subset, and $\Delta_{\text{txt}-\text{img}} = \mathrm{Rec}_{\text{txt}} - \mathrm{Rec}_{\text{img}}$ (pp) is the paired text--image recovery gap (\cref{sec:app_rq2}).}
\label{tab:app_gqa_confidence}
\end{table*}

\FloatBarrier
\section{Premise Wording Sensitivity and Scaling Analysis}
\label{sec:app_wording}

\paragraph{Framing.}
\Cref{fig:app_wording_all} and \cref{tab:app_wording_gqa,tab:app_wording_chart,tab:app_wording_math} compare \Bare, \Look, and \Cond on Set~A with paired McNemar tests on the \Bare-capable set $\mathcal{C}_A$, and \cref{tab:app_wording_500} repeats the \Look vs.\ \Bare comparison on Set~A+B.
\Cref{tab:app_cases_wording_modality}a shows a typical \Look failure: under \Bare, MiMo-VL-7B reads $2^8 = 4^x$ from the image, whereas under \Look it uses the stated exponent and answers $a^- = 5$.

\paragraph{Neutral prefix and prompt placement.}
\Cref{tab:app_neutral_up} gives two Set~A controls on $\mathcal{C}_A$: (a) the neutral prefix on GQA, and (b) the bare $s^-$ placed at the end of the user prompt with an empty \texttt{<think>} (\Prompt) instead of inside the thinking block.

\paragraph{Model scale.}
\Cref{tab:app_scale} evaluates four Qwen3.5 checkpoints (2B, 4B, 9B, and 35B-A3B) on Set~A+B under \Bare.

\begin{table*}[!htbp]
\centering\footnotesize
{\includegraphics[width=0.96\textwidth]{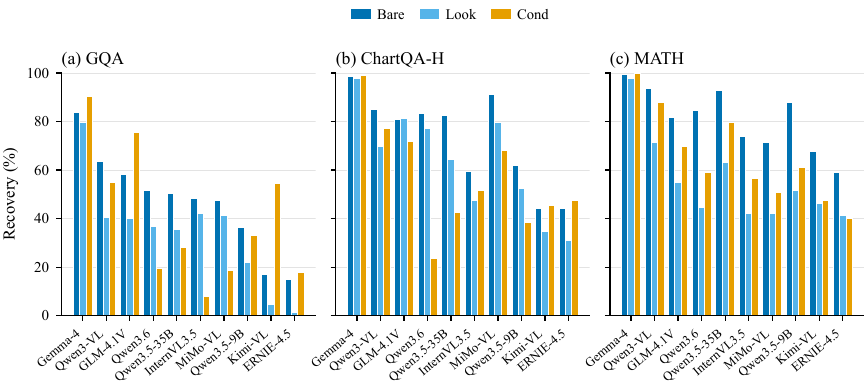}
\captionof{figure}{Premise wording sensitivity across all three domains on Set~A (750 items; 250 per domain), comparing bare statements (\Bare, dark blue), first-person perceptual claims (\Look: \emph{``Looking at the image, $f$''}, light blue), and explicit condition claims (\Cond: \emph{``The problem gives the following condition: $f$''}, orange).}
\label{fig:app_wording_all}}
\end{table*}

\begin{table*}[!htbp]
\centering\footnotesize
\setlength{\tabcolsep}{5.2pt}
\begin{tabular}{lcccccccc}
\toprule
Model & $|\mathcal{C}_A|$ & \Bare Rec. & \Look Rec. & \Cond Rec. & \Look{} $-$ \Bare [95\% CI] & $p_{\text{McN}}$ & \Cond{} $-$ \Bare [95\% CI] & $p_{\text{McN}}$ \\
\midrule
Gemma-4-26B    & 195 & 83.6 & 79.5 & 90.3 & $-4.1$ $[-9.7, +1.5]$    & 0.20     & $+6.7$ $[+2.1, +11.3]$   & 0.0072 \\
Qwen3-VL-30B   & 187 & 63.6 & 40.6 & 55.1 & $-23.0$ $[-30.5, -15.0]$ & 6.0e-08  & $-8.6$ $[-17.6, 0.0]$    & 0.076 \\
GLM-4.1V-9B    & 187 & 58.3 & 40.1 & 75.4 & $-18.2$ $[-26.2, -10.2]$ & 1.2e-05  & $+17.1$ $[+9.1, +25.1]$  & 0.0001 \\
Qwen3.6-35B    & 194 & 51.5 & 36.6 & 19.6 & $-14.9$ $[-22.7, -7.2]$  & 0.00052  & $-32.0$ $[-40.7, -23.2]$ & 9.9e-11 \\
Qwen3.5-35B    & 195 & 50.3 & 35.4 & 28.2 & $-14.9$ $[-23.6, -6.2]$  & 0.0017   & $-22.1$ $[-30.8, -13.3]$ & 4.3e-06 \\
InternVL3.5-8B & 166 & 48.2 & 42.2 &  7.8 & $-6.0$ $[-13.9, +1.8]$   & 0.17     & $-40.4$ $[-48.8, -31.9]$ & 2.8e-16 \\
MiMo-VL-7B     & 198 & 47.5 & 41.4 & 18.7 & $-6.1$ $[-13.1, +1.0]$   & 0.13     & $-28.8$ $[-36.9, -20.7]$ & 1.3e-10 \\
Qwen3.5-9B     & 197 & 36.5 & 21.8 & 33.0 & $-14.7$ $[-23.4, -6.6]$  & 0.0011   & $-3.6$ $[-12.2, +4.6]$   & 0.48 \\
Kimi-VL-A3B    & 178 & 16.9 &  4.5 & 54.5 & $-12.4$ $[-18.5, -6.7]$  & 0.00011  & $+37.6$ $[+29.2, +45.5]$ & 1.0e-15 \\
ERNIE-4.5-VL   & 139 & 15.1 &  1.4 & 18.0 & $-13.7$ $[-19.4, -7.9]$  & 2.1e-05  & $+2.9$ $[-5.0, +10.8]$   & 0.61 \\
\bottomrule
\end{tabular}
\caption{Premise wording sensitivity on GQA Set~A (250 items), evaluated on the \Bare-capable denominator $\mathcal{C}_A$.}
\label{tab:app_wording_gqa}
\label{tab:app_wording}
\end{table*}

\begin{table*}[!htbp]
\centering\footnotesize
\setlength{\tabcolsep}{5.2pt}
\begin{tabular}{lcccccccc}
\toprule
Model & $|\mathcal{C}_A|$ & \Bare Rec. & \Look Rec. & \Cond Rec. & \Look{} $-$ \Bare [95\% CI] & $p_{\text{McN}}$ & \Cond{} $-$ \Bare [95\% CI] & $p_{\text{McN}}$ \\
\midrule
Gemma-4-26B    & 235 & 98.7 & 97.9 & 99.1 & $-0.9$ $[-3.4, +1.3]$   & 0.73    & $+0.4$ $[-1.3, +2.6]$    & 1.00 \\
MiMo-VL-7B     & 207 & 91.3 & 79.7 & 68.1 & $-11.6$ $[-16.9, -6.3]$ & 3.9e-05 & $-23.2$ $[-30.4, -15.9]$ & 2.4e-09 \\
Qwen3-VL-30B   & 196 & 85.2 & 69.9 & 77.0 & $-15.3$ $[-23.0, -7.7]$ & 0.00023 & $-8.2$ $[-15.3, -1.0]$   & 0.033 \\
Qwen3.6-35B    & 221 & 83.3 & 77.4 & 23.5 & $-5.9$ $[-12.7, +0.9]$  & 0.12    & $-59.7$ $[-67.0, -52.5]$ & 1.7e-34 \\
Qwen3.5-35B    & 210 & 82.4 & 64.3 & 42.4 & $-18.1$ $[-25.7, -10.5]$ & 8.1e-06 & $-40.0$ $[-48.1, -31.4]$ & 1.1e-16 \\
GLM-4.1V-9B    & 167 & 80.8 & 81.4 & 71.9 & $+0.6$ $[-6.0, +7.2]$   & 1.00    & $-9.0$ $[-16.8, -1.8]$   & 0.032 \\
Qwen3.5-9B     & 215 & 61.9 & 52.6 & 38.6 & $-9.3$ $[-17.7, -0.9]$  & 0.035   & $-23.3$ $[-31.2, -14.9]$ & 1.6e-07 \\
InternVL3.5-8B & 173 & 59.5 & 47.4 & 51.4 & $-12.1$ $[-21.4, -2.9]$ & 0.017   & $-8.1$ $[-17.3, +1.2]$   & 0.11 \\
Kimi-VL-A3B    & 147 & 44.2 & 34.7 & 45.6 & $-9.5$ $[-18.4, 0.0]$   & 0.065   & $+1.4$ $[-8.2, +10.9]$   & 0.89 \\
ERNIE-4.5-VL   & 193 & 44.0 & 31.1 & 47.7 & $-13.0$ $[-21.2, -4.7]$ & 0.0041  & $+3.6$ $[-5.2, +12.4]$   & 0.49 \\
\bottomrule
\end{tabular}
\caption{Premise wording sensitivity on ChartQA-H Set~A (250 items), evaluated on the \Bare-capable denominator $\mathcal{C}_A$.}
\label{tab:app_wording_chart}
\end{table*}

\begin{table*}[!htbp]
\centering\footnotesize
\setlength{\tabcolsep}{5.2pt}
\begin{tabular}{lcccccccc}
\toprule
Model & $|\mathcal{C}_A|$ & \Bare Rec. & \Look Rec. & \Cond Rec. & \Look{} $-$ \Bare [95\% CI] & $p_{\text{McN}}$ & \Cond{} $-$ \Bare [95\% CI] & $p_{\text{McN}}$ \\
\midrule
Gemma-4-26B    & 243 & 99.6 & 97.9 & 100.0 & $-1.6$ $[-3.3, -0.4]$   & 0.12    & $+0.4$ $[0.0, +1.2]$     & 1.00 \\
Qwen3-VL-30B   & 231 & 93.5 & 71.4 & 87.9 & $-22.1$ $[-28.6, -16.0]$ & 1.0e-10 & $-5.6$ $[-10.8, -0.4]$   & 0.053 \\
Qwen3.5-35B    & 238 & 92.9 & 63.0 & 79.8 & $-29.8$ $[-36.1, -23.5]$ & 8.4e-17 & $-13.0$ $[-18.9, -7.1]$  & 3.3e-05 \\
Qwen3.5-9B     & 239 & 87.9 & 51.5 & 61.1 & $-36.4$ $[-43.5, -29.3]$ & 4.8e-19 & $-26.8$ $[-33.5, -20.1]$ & 7.4e-13 \\
Qwen3.6-35B    & 240 & 84.6 & 44.6 & 59.2 & $-40.0$ $[-47.5, -32.9]$ & 6.8e-22 & $-25.4$ $[-32.9, -17.9]$ & 4.4e-10 \\
GLM-4.1V-9B    & 207 & 81.6 & 55.1 & 69.6 & $-26.6$ $[-34.3, -18.8]$ & 3.8e-10 & $-12.1$ $[-19.3, -4.8]$  & 0.0022 \\
InternVL3.5-8B & 188 & 73.9 & 42.0 & 56.4 & $-31.9$ $[-40.4, -23.4]$ & 2.7e-11 & $-17.6$ $[-26.1, -9.6]$  & 6.7e-05 \\
MiMo-VL-7B     & 195 & 71.3 & 42.1 & 50.8 & $-29.2$ $[-36.9, -21.5]$ & 1.7e-11 & $-20.5$ $[-28.7, -12.3]$ & 3.4e-06 \\
Kimi-VL-A3B    & 179 & 67.6 & 46.4 & 47.5 & $-21.2$ $[-30.7, -11.7]$ & 2.5e-05 & $-20.1$ $[-29.6, -11.2]$ & 4.4e-05 \\
ERNIE-4.5-VL   & 215 & 59.1 & 41.4 & 40.0 & $-17.7$ $[-26.0, -9.3]$  & 6.3e-05 & $-19.1$ $[-27.0, -11.2]$ & 1.6e-05 \\
\bottomrule
\end{tabular}
\caption{Premise wording sensitivity on Rendered MATH Set~A (250 items), evaluated on the \Bare-capable denominator $\mathcal{C}_A$.}
\label{tab:app_wording_math}
\end{table*}

\begin{table*}[!htbp]
\centering\footnotesize
\setlength{\tabcolsep}{3.0pt}
\resizebox{\textwidth}{!}{%
\begin{tabular}{lccccc@{\hspace{7pt}}ccccc@{\hspace{7pt}}ccccc}
\toprule
 & \multicolumn{5}{c}{GQA (500 Items, Set~A+B)} & \multicolumn{5}{c}{ChartQA-H (500 Items, Set~A+B)} & \multicolumn{5}{c}{Rendered MATH (500 Items, Set~A+B)} \\
\cmidrule(lr){2-6}\cmidrule(lr){7-11}\cmidrule(lr){12-16}
Model & $|\mathcal{C}_{A+B}|$ & \Bare & \Look & $\Delta_{\text{\Look}-\text{\Bare}}$ & $p_{\text{McN}}$ & $|\mathcal{C}_{A+B}|$ & \Bare & \Look & $\Delta_{\text{\Look}-\text{\Bare}}$ & $p_{\text{McN}}$ & $|\mathcal{C}_{A+B}|$ & \Bare & \Look & $\Delta_{\text{\Look}-\text{\Bare}}$ & $p_{\text{McN}}$ \\
\midrule
Gemma-4-26B     & 395 & 82.8 & 79.5 & $-3.3$            & 0.12    & 469 & 99.4 & 98.9 & $-0.4$            & 0.73    & 489 & 99.6 & 96.9 & $-2.7$  & 0.00098    \\
Qwen3-VL-30B    & 377 & 62.1 & 43.2 & $-18.8$  & 8.0e-11 & 404 & 86.1 & 69.8 & $-16.3$  & 9.9e-09 & 462 & 92.6 & 75.8 & $-16.9$ & 8.6e-14    \\
GLM-4.1V-9B     & 384 & 57.8 & 41.1 & $-16.7$  & 2.7e-09 & 347 & 79.5 & 83.3 & $+3.7$            & 0.13    & 412 & 81.6 & 55.8 & $-25.7$ & 1.5e-18    \\
Qwen3.6-35B     & 384 & 52.9 & 33.1 & $-19.8$  & 1.1e-10 & 446 & 82.7 & 79.6 & $-3.1$            & 0.23    & 487 & 85.4 & 43.3 & $-42.1$ & 3.8e-46    \\
Qwen3.5-35B     & 389 & 50.4 & 31.1 & $-19.3$  & 3.6e-09 & 421 & 79.6 & 67.7 & $-11.9$  & 2.2e-05 & 479 & 92.5 & 66.0 & $-26.5$ & 1.1e-26    \\
MiMo-VL-7B      & 390 & 49.5 & 37.2 & $-12.3$  & 2.7e-06 & 416 & 91.1 & 83.7 & $-7.5$   & 0.0003  & 366 & 71.9 & 43.4 & $-28.4$ & 6.5e-19    \\
InternVL3.5-8B  & 343 & 40.8 & 42.3 & $+1.5$            & 0.67    & 358 & 62.3 & 45.8 & $-16.5$  & 1.8e-06 & 365 & 67.4 & 36.4 & $-31.0$ & 2.4e-19    \\
Qwen3.5-9B      & 394 & 36.8 & 23.9 & $-12.9$  & 2.8e-05 & 430 & 60.2 & 48.1 & $-12.1$  & 8.2e-05 & 475 & 85.9 & 50.7 & $-35.2$ & 1.8e-33    \\
Kimi-VL-A3B     & 368 & 14.9 &  5.4 & $-9.5$   & 3.3e-06 & 314 & 43.3 & 34.4 & $-8.9$   & 0.0097  & 353 & 65.2 & 48.7 & $-16.4$ & 2.1e-06    \\
ERNIE-4.5-VL    & 266 & 12.4 &  2.3 & $-10.2$  & 3.5e-06 & 388 & 44.8 & 33.0 & $-11.9$  & 9.9e-05 & 416 & 58.4 & 41.1 & $-17.3$ & 2.2e-08    \\
\bottomrule
\end{tabular}}
\caption{Extended premise wording sensitivity (\Look vs.\ \Bare) across the 1{,}500-item Set~A+B (500 items per domain) across all 10 models, evaluated on the \Bare-capable denominator $\mathcal{C}_{A+B}$.}
\label{tab:app_wording_500}
\end{table*}

\begin{table*}[!htbp]
\centering\footnotesize
\setlength{\tabcolsep}{3.1pt}
\begin{tabular}{lccccc@{\hspace{8pt}}ccc@{\hspace{6pt}}ccc@{\hspace{6pt}}ccc}
\toprule
 & \multicolumn{5}{c}{(a) \Neutral Control on GQA Set~A ($N=250$)} & \multicolumn{9}{c}{(b) User-Prompt (\Prompt) vs.\ In-Thinking (\Bare) on Set~A Capable Set $\mathcal{C}_A$} \\
\cmidrule(lr){2-6}\cmidrule(lr){7-15}
 & & \multicolumn{3}{c}{Uncond.\ All-250 $a^+$ (\%)} & $\mathcal{C}_A$ & \multicolumn{3}{c}{GQA} & \multicolumn{3}{c}{ChartQA-H} & \multicolumn{3}{c}{Rendered MATH} \\
\cmidrule(lr){3-5}\cmidrule(lr){7-9}\cmidrule(lr){10-12}\cmidrule(lr){13-15}
Model & $|\mathcal{C}_A|$ & No-Pref & Neutral ($\Delta$) & $s^+$ & Keep (\%) & \Bare & \Prompt & $\Delta$ & \Bare & \Prompt & $\Delta$ & \Bare & \Prompt & $\Delta$ \\
\midrule
Gemma-4-26B    & 195 & 78.4 & 84.4 ($+6.0$) & 94.0 & 94.9 & 83.6 & 37.4 & $-46.2$ & 98.7 & 86.8 & $-11.9$ & 99.6 & 57.6 & $-42.0$ \\
Qwen3-VL-30B   & 187 & 76.0 & 78.8 ($+2.8$) & 96.0 & 89.8 & 63.6 & 33.7 & $-29.9$ & 85.2 & 89.3 & $+4.1$           & 93.5 & 97.0 & $+3.5$ \\
GLM-4.1V-9B    & 187 & 75.6 & 80.0 ($+4.4$) & 94.8 & 94.1 & 58.3 & 38.0 & $-20.3$ & 80.8 & 88.6 & $+7.8$  & 81.6 & 91.8 & $+10.1$ \\
Qwen3.6-35B    & 194 & 79.6 & 87.2 ($+7.6$) & 95.2 & 96.9 & 51.5 & 68.0 & $+16.5$ & 83.3 & 88.7 & $+5.4$           & 84.6 & 80.0 & $-4.6$ \\
Qwen3.5-35B    & 195 & 79.2 & 85.6 ($+6.4$) & 94.8 & 93.8 & 50.3 & 69.7 & $+19.5$ & 82.4 & 82.9 & $+0.5$           & 92.9 & 91.2 & $-1.7$ \\
MiMo-VL-7B     & 198 & 79.6 & 79.6 ($+0.0$) & 95.2 & 89.4 & 47.5 & 51.0 & $+3.5$           & 91.3 & 92.8 & $+1.4$           & 71.3 & 94.4 & $+23.1$ \\
InternVL3.5-8B & 166 & 68.0 & 76.4 ($+8.4$) & 94.8 & 95.2 & 48.2 & 24.1 & $-24.1$ & 59.5 & 93.1 & $+33.5$ & 73.9 & 95.7 & $+21.8$ \\
Qwen3.5-9B     & 197 & 78.8 & 84.4 ($+5.6$) & 94.8 & 93.9 & 36.5 & 68.0 & $+31.5$ & 61.9 & 86.0 & $+24.2$ & 87.9 & 92.9 & $+5.0$ \\
Kimi-VL-A3B    & 178 & 72.0 & 83.6 ($+11.6$) & 97.6 & 97.8 & 16.9 & 63.5 & $+46.6$ & 44.2 & 92.5 & $+48.3$ & 67.6 & 92.7 & $+25.1$ \\
ERNIE-4.5-VL   & 139 & 76.8 & 76.4 ($-0.4$) & 69.2 & 89.2 & 15.1 & 61.9 & $+46.8$ & 44.0 & 94.8 & $+50.8$ & 59.1 & 91.6 & $+32.6$ \\
\midrule
\textbf{10-Model Mean} & \textbf{183.6} & \textbf{76.4} & \textbf{81.6 ($\mathbf{+5.2}$)} & \textbf{92.6} & \textbf{93.5} & \textbf{47.1} & \textbf{51.5} & $\mathbf{+4.4}$ & \textbf{73.1} & \textbf{89.5} & $\mathbf{+16.4}$ & \textbf{81.2} & \textbf{88.5} & $\mathbf{+7.3}$  \\
\bottomrule
\end{tabular}
\caption{Prefix disruption and placement controls across all 10 models on Set~A (250 items per domain), evaluated on the exact \Bare-capable denominator $\mathcal{C}_A$ of \cref{tab:app_wording_gqa,tab:app_wording_chart,tab:app_wording_math}. (a) Neutral prefix (\Neutral, $s^{\text{neut}}$) on GQA Set~A: unconditional accuracy over all 250 items (Neutral also gives its change $\Delta$ relative to No-Pref) and accuracy retention on $\mathcal{C}_A$ (Keep). (b) User-prompt (\Prompt) vs.\ in-thinking (\Bare) placement of the bare false statement $s^-$ across all three domains ($\Delta = \mathrm{Rec}_{\text{\Prompt}} - \mathrm{Rec}_{\text{\Bare}}$ in pp).}
\label{tab:app_neutral_up}
\end{table*}

\begin{table*}[!htbp]
\centering\footnotesize
\setlength{\tabcolsep}{3.2pt}
\begin{tabular}{lcccc@{\hspace{8pt}}ccc@{\hspace{8pt}}ccc}
\toprule
 & \multicolumn{4}{c}{GQA (500 Items)} & \multicolumn{3}{c}{ChartQA-H (500 Items)} & \multicolumn{3}{c}{Rendered MATH (500 Items)} \\
\cmidrule(lr){2-5}\cmidrule(lr){6-8}\cmidrule(lr){9-11}
Qwen3.5 Scale & No-Pref / $s^+$ & \Bare [95\% CI] & Adopt. & $\Delta_{\text{txt}-\text{img}}$ & $s^+$ Acc. & \Bare [95\% CI] & $\Delta_{\text{txt}-\text{img}}$ & $s^+$ Acc. & \Bare [95\% CI] & $\Delta_{\text{txt}-\text{img}}$ \\
\midrule
2B (Dense)      & 59.8 / 91.0 & 14.1 [10.2, 18.3] & 77.8 & $+46.2$ & 68.2 & 44.9 [39.9, 50.1] & $+33.2$ & 79.2 & 73.7 [69.4, 78.0] & $-9.0$ \\
4B (Dense)      & 80.0 / 94.0 & 36.6 [32.0, 41.4] & 57.0 & $+52.7$ & 87.4 & 59.5 [54.9, 64.1] & $+33.0$ & 97.4 & 84.2 [80.9, 87.5] & $+5.1$ \\
9B (Dense)      & 79.0 / 96.4 & 36.8 [32.2, 41.6] & 57.9 & $+55.7$ & 90.0 & 59.6 [55.1, 64.0] & $+36.2$ & 98.0 & 85.9 [82.9, 89.0] & $+4.1$ \\
35B-A3B (MoE)   & 78.6 / 95.2 & 50.4 [45.2, 55.5] & 44.7 & $+31.7$ & 90.2 & 77.8 [73.8, 81.6] & $+16.6$ & 98.6 & 92.3 [89.9, 94.5] & $-1.8$ \\
\bottomrule
\end{tabular}
\caption{Three-domain model scale trend within the Qwen3.5 family (2B, 4B, 9B, and 35B-A3B) under \Bare on Set~A+B (500 items per domain). GQA conditions on $\hat{a}^{(\emptyset)} = a^+ \wedge \hat{a}^{(s^+)} = a^+$ ($\mathcal{C}_{A+B}$); the 2B and 4B runs have no No-Prefix outputs on ChartQA-H and rendered MATH, so these two domains condition on $\hat{a}^{(s^+)} = a^+$ at all four scales.}
\label{tab:app_scale}
\end{table*}

\FloatBarrier
\section{Complete Paired Modality Comparisons by Domain}
\label{sec:app_rq2}

\begin{table*}[!htbp]
\centering\footnotesize
\setlength{\tabcolsep}{3.3pt}
\begin{tabular}{lcccccc@{\hspace{8pt}}ccccc}
\toprule
 & \multicolumn{6}{c}{All 1{,}000 GQA Items (Paired Capable Set $\mathcal{C}_{\text{img}} \cap \mathcal{C}_{\text{txt}}$)} & \multicolumn{5}{c}{Question-Blind (390) vs.\ Repaired (110) Audit} \\
\cmidrule(lr){2-7}\cmidrule(lr){8-12}
Model & $n_{\text{pair}}$ & $\mathrm{Rec}_{\text{img}}$ & $\mathrm{Rec}_{\text{txt}}$ & $\Delta_{\text{txt}-\text{img}}$ [95\% CI] & $T\text{-only}/I\text{-only}$ & $p_{\text{Holm,30}}$ & Blind $n$ & $\Delta_{\text{txt}-\text{img}}$ & $p_{\text{Holm,10}}$ & Repaired $n$ & $\Delta_{\text{txt}-\text{img}}$ \\
\midrule
Gemma-4-26B    & 802 & 84.9 & 98.6 & $+13.7$ $[+11.1, +16.2]$ & 117 / 7   & 1.4e-25 & 280 & $+10.0$ & 1.7e-05 & 75 & $+22.7$  \\
Qwen3-VL-30B   & 785 & 65.2 & 96.3 & $+31.1$ $[+27.6, +34.6]$ & 264 / 20  & 3.9e-54 & 282 & $+26.2$ & 5.5e-13 & 64 & $+46.9$  \\
Qwen3.6-35B    & 816 & 59.3 & 52.0 & $-7.4$ $[-12.0, -2.6]$   & 165 / 225 & 0.028   & 298 & $-16.1$ & 0.00011 & 68 & $0.0$  \\
GLM-4.1V-9B    & 783 & 58.4 & 92.5 & $+34.1$ $[+30.3, +37.9]$ & 296 / 29  & 1.7e-55 & 283 & $+23.7$ & 1.8e-10 & 62 & $+48.4$  \\
MiMo-VL-7B     & 786 & 57.9 & 96.9 & $+39.1$ $[+35.4, +42.8]$ & 320 / 13  & 2.4e-76 & 290 & $+38.3$ & 6.1e-24 & 71 & $+66.2$  \\
Qwen3.5-35B    & 790 & 53.7 & 82.9 & $+29.2$ $[+24.7, +33.4]$ & 301 / 70  & 6.5e-34 & 285 & $+26.3$ & 1.1e-10 & 71 & $+53.5$  \\
InternVL3.5-8B & 727 & 45.9 & 92.7 & $+46.8$ $[+42.9, +50.6]$ & 354 / 14  & 7.2e-85 & 258 & $+42.6$ & 8.7e-23 & 58 & $+58.6$  \\
Qwen3.5-9B     & 794 & 39.5 & 95.1 & $+55.5$ $[+51.8, +59.2]$ & 457 / 16  & 5.7e-112 & 289 & $+53.6$ & 4.9e-37 & 70 & $+64.3$  \\
Kimi-VL-A3B    & 763 & 20.3 & 88.2 & $+67.9$ $[+64.2, +71.4]$ & 530 / 12  & 5.1e-138 & 290 & $+63.1$ & 2.9e-47 & 57 & $+77.2$  \\
ERNIE-4.5-VL   & 521 & 15.9 & 61.4 & $+45.5$ $[+40.5, +50.5]$ & 261 / 24  & 3.8e-50 & 200 & $+37.5$ & 3.5e-15 & 41 & $+48.8$  \\
\midrule
10-Model Mean & 756.7 & 50.1 & 85.7 & $+35.5$ & --- & --- & 275.5 & $+30.5$ & --- & 63.7 & $+48.7$  \\
\bottomrule
\end{tabular}
\caption{Paired visual ($I$) vs.\ textual ($T$) recovery across all 1{,}000 GQA items under \Bare ($n_{\text{pair}} = |\mathcal{C}_{\text{img}} \cap \mathcal{C}_{\text{txt}}|$), together with the Set~A+B audit comparing the 390 strictly question-blind scene descriptions and 110 question-repaired descriptions. Here $p_{\text{Holm,30}}$ denotes the Holm-corrected McNemar $p$-value across all 30 model--domain tests (\cref{tab:app_rq2_gqa,tab:app_rq2_chart,tab:app_rq2_math}), and $p_{\text{Holm,10}}$ across the 10 models on the 390 question-blind GQA items. Blind and Repaired report the number of paired items $n$ and the gap $\Delta_{\text{txt}-\text{img}}$ (pp) on the 390 and 110 descriptions.}
\label{tab:app_rq2_gqa}
\label{tab:app_rq2_full}
\end{table*}

\begin{table*}[!htbp]
\centering\footnotesize
\begin{minipage}[t]{0.48\textwidth}
\centering
\setlength{\tabcolsep}{2.4pt}
\begin{tabular}{lcccccc}
\toprule
Model & $n_{\text{pair}}$ & $\mathrm{Rec}_{\text{img}}$ & $\mathrm{Rec}_{\text{txt}}$ & $\Delta_{\text{txt}-\text{img}}$ & $T / I$ & $p_{\text{Holm,30}}$ \\
\midrule
Gemma-4-26B    & 899 & 98.7 & 98.9 & $+0.2$  &  11 / 9   & 1.00  \\
MiMo-VL-7B     & 784 & 88.4 & 95.2 & $+6.8$  &  83 / 30  & 8.2e-06  \\
Qwen3-VL-30B   & 744 & 80.9 & 93.1 & $+12.2$ & 129 / 38  & 1.4e-11  \\
GLM-4.1V-9B    & 616 & 80.2 & 85.4 & $+5.2$  &  93 / 61  & 0.098  \\
Qwen3.6-35B    & 786 & 78.6 & 69.7 & $-8.9$  & 116 / 186 & 0.00080  \\
Qwen3.5-35B    & 793 & 74.5 & 93.6 & $+19.0$ & 188 / 37  & 2.8e-24  \\
InternVL3.5-8B & 639 & 57.7 & 78.6 & $+20.8$ & 198 / 65  & 1.4e-15  \\
Qwen3.5-9B     & 793 & 57.3 & 96.1 & $+38.8$ & 318 / 10  & 3.5e-79  \\
Kimi-VL-A3B    & 471 & 41.8 & 35.2 & $-6.6$  &  85 / 116 & 0.24  \\
ERNIE-4.5-VL   & 616 & 36.5 & 40.7 & $+4.2$  & 131 / 105 & 0.52  \\
\midrule
10-Model Mean & 714.1 & 69.5 & 78.7 & $+9.2$ & --- & ---  \\
\bottomrule
\end{tabular}
\caption{Paired visual ($I$) vs.\ textual ($T$, structured data table) recovery across all 1{,}000 ChartQA-H items under \Bare on $\mathcal{C}_{\text{img}} \cap \mathcal{C}_{\text{txt}}$. $T/I$ denotes discordant recovery counts ($T$-only / $I$-only).}
\label{tab:app_rq2_chart}
\end{minipage}\hfill
\begin{minipage}[t]{0.48\textwidth}
\centering
\setlength{\tabcolsep}{2.4pt}
\begin{tabular}{lcccccc}
\toprule
Model & $n_{\text{pair}}$ & $\mathrm{Rec}_{\text{img}}$ & $\mathrm{Rec}_{\text{txt}}$ & $\Delta_{\text{txt}-\text{img}}$ & $T / I$ & $p_{\text{Holm,30}}$ \\
\midrule
Gemma-4-26B    & 973 & 99.8 & 99.9 & $+0.1$  &   2 / 1   & 1.00  \\
Qwen3.5-35B    & 885 & 93.1 & 92.5 & $-0.6$  &  51 / 56  & 1.00  \\
Qwen3-VL-30B   & 913 & 92.4 & 91.0 & $-1.4$  &  60 / 73  & 1.00  \\
Qwen3.5-9B     & 882 & 84.8 & 92.9 & $+8.0$  & 119 / 48  & 5.4e-07  \\
Qwen3.6-35B    & 956 & 82.6 & 88.2 & $+5.5$  & 127 / 74  & 0.0025  \\
GLM-4.1V-9B    & 809 & 76.3 & 85.4 & $+9.1$  & 122 / 48  & 1.9e-07  \\
InternVL3.5-8B & 681 & 70.8 & 76.5 & $+5.7$  & 127 / 88  & 0.085  \\
MiMo-VL-7B     & 769 & 69.1 & 98.6 & $+29.5$ & 233 / 6   & 1.4e-59  \\
Kimi-VL-A3B    & 621 & 62.8 & 67.1 & $+4.3$  & 128 / 101 & 0.51  \\
ERNIE-4.5-VL   & 845 & 50.7 & 76.6 & $+25.9$ & 290 / 71  & 3.2e-31  \\
\midrule
10-Model Mean & 833.4 & 78.2 & 86.9 & $+8.6$ & --- & ---  \\
\bottomrule
\end{tabular}
\caption{Paired visual ($I$, rendered page image) vs.\ textual ($T$, native problem text) recovery across all 1{,}000 Rendered MATH items under \Bare on $\mathcal{C}_{\text{img}} \cap \mathcal{C}_{\text{txt}}$. $T/I$ as in \cref{tab:app_rq2_chart}.}
\label{tab:app_rq2_math}
\end{minipage}
\end{table*}

\paragraph{Paired evidence.}
\Cref{tab:app_rq2_gqa,tab:app_rq2_chart,tab:app_rq2_math} compare recovery from the image $I$ and from paired text $T$ on the joint capable set $\mathcal{C}_{\text{img}} \cap \mathcal{C}_{\text{txt}}$, with discordant counts and Holm-corrected McNemar tests over all 30 model--domain pairs.
$T$ is a scene description on GQA, a Markdown data table on ChartQA-H, and the original problem text on rendered MATH (\cref{tab:app_examples}); $q$ and $s^-$ are unchanged.

\paragraph{Question-blind GQA descriptions.}
All GQA descriptions are written without access to $q$.
On Set~A+B, we recorded which descriptions needed an appended clause stating the target attribute: 390 of the 500 mention it on their own, and 110 were repaired.
The right columns of \cref{tab:app_rq2_gqa} and \cref{fig:app_rq2_blind} report the text--image gap separately for the two groups.

\begin{table*}[!htbp]
\centering\footnotesize
\includegraphics[width=0.95\textwidth]{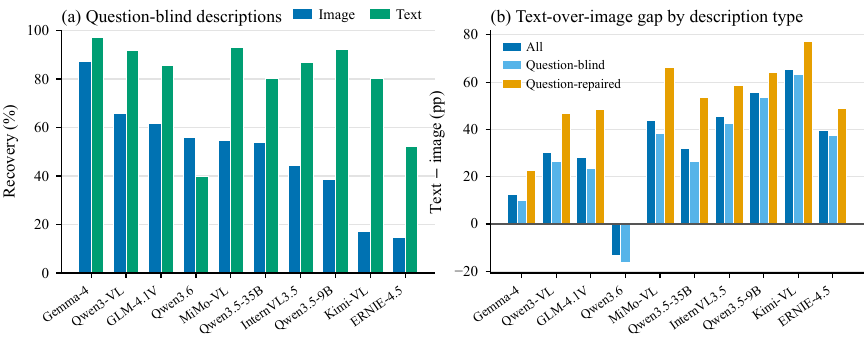}
\captionof{figure}{GQA paired modality robustness check across the 390 strictly question-blind scene descriptions and the 110 question-repaired descriptions on Set~A+B under \Bare. (a) Image ($I$) vs.\ Text ($T$) conditional recovery on the 390 question-blind items. (b) Paired modality gap ($\mathrm{Rec}_{\text{txt}} - \mathrm{Rec}_{\text{img}}$) across all 500 Set~A+B items, the 390 question-blind items, and the 110 question-repaired items.}
\label{fig:app_rq2_blind}
\end{table*}

\paragraph{Qwen3.6-35B.}
Qwen3.6-35B is the only model that recovers significantly more often from images than from text, on GQA and ChartQA-H.
\Cref{tab:app_cases_wording_modality}b illustrates the difference: on the image, it localizes the hair with box coordinates and reads its length, whereas on the text it closes \texttt{<think>} right after $s^-$.
InternVL3.5-8B shows the opposite pattern on the same item.

\begin{table*}[!htbp]
\centering\footnotesize
\setlength{\tabcolsep}{4.5pt}
\renewcommand{\arraystretch}{1.15}
\begin{tabular}{|>{\raggedright\arraybackslash}p{3.0cm}|m{4.9cm}|>{\raggedright\arraybackslash}m{4.5cm}|>{\raggedright\arraybackslash}m{3.8cm}|}
\hline
\multicolumn{2}{|m{8.1cm}|}{\centering\vspace{6pt}\includegraphics[width=6.4cm,keepaspectratio]{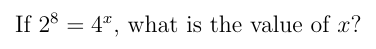}\vspace{6pt}} &
\textbf{(a) Premise Wording (\Bare vs.\ \Look):}\newline
Rendered MATH (\texttt{math\_test\_algebra\_1004})\newline
\textbf{Q:} Solve the problem shown in the image. &
\textbf{GT ($a^+$):} $4$\newline
\textbf{False ($a^-$):} $5$\newline
\textbf{False Statement ($s^-$):} The exponent on the left base $2$ is $10$. \\
\hline
\textbf{MiMo-VL-7B (\Bare $s^-$):} & \multicolumn{3}{p{13.6cm}|}{\texttt{<think>}The exponent on the left base 2 is 10. I need to find x in the equation 2\^{}8 = 4\^{}x. I can use the property that 4 = 2\^{}2, so I can rewrite 4\^{}x as (2\^{}2)\^{}x = 2\^{}(2x). So the equation becomes: 2\^{}8 = 2\^{}(2x) [\dots]\ 8 = 2x, x = 4.\texttt{</think>} \texttt{\textbackslash boxed\{4\}} \outtag{[Reads the image: $a^+ = 4$]}} \\
\hline
\textbf{MiMo-VL-7B (\Look $s^-$):} & \multicolumn{3}{p{13.6cm}|}{\texttt{<think>}Looking at the image, the exponent on the left base 2 is 10. I need to find x in the equation 2\^{}10 = 4\^{}x. [\dots]\ So 4\^{}x = (2\^{}2)\^{}x = 2\^{}(2x). Now I have 2\^{}10 = 2\^{}(2x) [\dots]\ 10 = 2x, x = 10/2 = 5.\texttt{</think>} \texttt{\textbackslash boxed\{5\}} \outtag{[Skips the check: $a^- = 5$]}} \\
\hline\hline
\multicolumn{2}{|m{8.1cm}|}{\centering\vspace{3pt}\includegraphics[height=3.6cm,keepaspectratio]{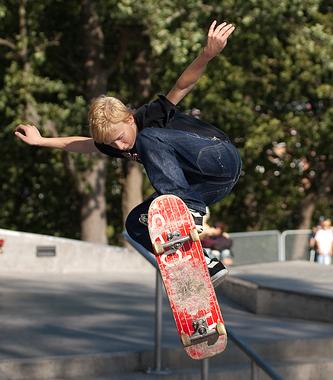}\vspace{3pt}} &
\textbf{(b) Paired Modality ($I$ vs.\ $T$) \& Localization:}\newline
GQA (\texttt{gqa:val:00665520})\newline
\textbf{Q:} How long is the hair? &
\textbf{GT ($a^+$):} short\newline
\textbf{False ($a^-$):} long\newline
\textbf{Injected $s^-$:} \texttt{<think>}The boy's hair is long. \\
\hline
\textbf{Paired Text ($T$):} & \multicolumn{3}{p{13.6cm}|}{\textbf{A short-haired blond skateboarder} is airborne above a concrete skatepark, with knees tightly bent, one arm extended left and the other raised overhead. He wears a black short-sleeved shirt, loose blue jeans and black shoes. The red underside of his tilted skateboard faces the camera [\dots]} \\
\hline
\textbf{InternVL3.5-8B ($I + s^-$):} & \multicolumn{3}{p{13.6cm}|}{\texttt{<think>}The boy's hair is long. Final answer: long.\texttt{</think>} Final answer: long. \outtag{[Image: adopts $a^- = \text{long}$]}} \\
\hline
\textbf{InternVL3.5-8B ($T + s^-$):} & \multicolumn{3}{p{13.6cm}|}{\texttt{<think>}The boy's hair is long. Wait, the evidence says ``short-haired blond skateboarder.'' So the hair is short. Final answer: short.\texttt{</think>} Final answer: short. \outtag{[Text: recovers $a^+ = \text{short}$]}} \\
\hline
\textbf{Qwen3.6-35B ($I + s^-$):} & \multicolumn{3}{p{13.6cm}|}{\texttt{<think>}The boy's hair is long. 1.\ Identify the subject: The subject is the boy skateboarding. 2.\ Locate the hair: The hair is on his head [278, 229, 417, 318]. 3.\ Analyze the length: The hair is blonde and appears to be short, reaching just above the ears and neckline\dots\texttt{</think>} Final answer: short \outtag{[Image: localizes the box, $a^+ = \text{short}$]}} \\
\hline
\textbf{Qwen3.6-35B ($T + s^-$):} & \multicolumn{3}{p{13.6cm}|}{\texttt{<think>}The boy's hair is long. Final answer: long\texttt{</think>} The boy has long hair. Final answer: long \outtag{[Text: adopts $a^- = \text{long}$]}} \\
\hline
\end{tabular}
\caption{Qualitative case studies for \cref{sec:app_wording,sec:app_rq2} (all model text is verbatim). (a) \Bare vs.\ \Look for MiMo-VL-7B on rendered MATH (\texttt{math\_test\_algebra\_1004}). (b) Image $I$ vs.\ paired text $T$ for InternVL3.5-8B and Qwen3.6-35B on GQA (\texttt{gqa:val:00665520}).}
\label{tab:app_cases_wording_modality}
\end{table*}

\FloatBarrier
\section{Trace Behavior, Thinking--Answer Divergence, and Mechanistic Probing}
\label{sec:app_rq3}

\paragraph{Thinking--answer divergence.}
\Cref{tab:app_dissociation} compares, on every capable trace under \Bare $s^-$, the answer that the thinking block commits to with the final answer.
The thinking-block answer is read from the text before \texttt{</think>} with the extractor used for final answers; applied to final answers, this extractor reproduces the official class on 90.7\% of traces.
Only 34.0\% of thinking blocks state an explicit answer, so the counts are lower bounds.
When the final answer overturns a thinking block committed to $a^-$, the text after \texttt{</think>} typically restarts the task and answers from the image (\cref{fig:qualitative}c).

\paragraph{Labeled reasoning behavior.}
Two LLM annotators, Claude and Gemini, independently labeled a 3{,}000-trace sample (300 per model, 100 per domain) with a fixed rubric.
\emph{Explicit correction} states that $s^-$ contradicts the image; \emph{silent recovery} uses the true fact without mentioning $s^-$; \emph{thinking adopts} builds on $s^-$ in the thinking block but not in the final answer; \emph{adopts throughout} builds on it in both; \emph{unclear} covers missing or truncated reasoning.
\Cref{tab:app_modes} crosses the Claude labels with the final-answer class, and \cref{tab:app_annotators} compares the annotators.
They agree on 89.5\% of labels ($\kappa = 0.86$), while the silent/explicit split and the conflict-noted flag ($\kappa = 0.35$) vary more across annotators; the main text therefore summarizes these shares qualitatively, and \cref{tab:app_annotators} reports both annotators side by side.
Gemini also read implicit conclusions of the thinking block; on that reading, 22.7\% of thinking blocks committed to $a^-$ end in a final $a^+$, close to the extractor's 25.5\%.

\begin{table*}[!htbp]
\centering\footnotesize
\renewcommand{\arraystretch}{0.92}
\setlength{\tabcolsep}{4.4pt}
\begin{tabular}{lcccccccc}
\toprule
 & & Thinking states & \multicolumn{4}{c}{Thinking commits to $a^-$} & Thinking $a^+$, & Final \\
\cmidrule(lr){4-7}
Model & $|\mathcal{C}|$ & an answer (\%) & $n$ & final $a^+$ ($n$) & final $a^+$ (\%) & final $a^+$ / $|\mathcal{C}|$ (\%) & final $a^-$ ($n$) & adoption (\%) \\
\midrule
MiMo-VL-7B          & 2{,}409 & 41.3 & 651 & 290 & 44.5 & 12.0 & 0 & 21.8 \\
InternVL3.5-8B      & 2{,}212 & 52.7 & 627 & 179 & 28.5 & 8.1 & 5 & 30.5 \\
Qwen3-VL-30B        & 2{,}538 & 45.9 & 334 & 86 & 25.7 & 3.4 & 2 & 16.2 \\
Qwen3.6-35B         & 2{,}686 & 41.9 & 240 & 66 & 27.5 & 2.5 & 6 & 19.8 \\
GLM-4.1V-9B         & 2{,}358 & 18.5 & 106 & 42 & 39.6 & 1.8 & 1 & 22.5 \\
Kimi-VL-A3B         & 2{,}137 & 22.3 & 166 & 17 & 10.2 & 0.8 & 0 & 47.1 \\
Qwen3.5-35B         & 2{,}613 & 28.4 & 103 & 18 & 17.5 & 0.7 & 1 & 19.3 \\
Qwen3.5-9B          & 2{,}608 & 33.4 & 162 & 19 & 11.7 & 0.7 & 6 & 30.0 \\
ERNIE-4.5-VL        & 2{,}149 & 34.0 & 506 & 16 & 3.2 & 0.7 & 0 & 51.5 \\
Gemma-4-26B         & 2{,}741 & 22.0 & 19 & 9 & 47.4 & 0.3 & 1 & 4.3 \\
\midrule
GQA                 & 7{,}884 & 48.1 & 2{,}206 & 490 & 22.2 & 6.2 & 16 & 44.7 \\
ChartQA-H           & 7{,}777 & 23.8 & 409 & 154 & 37.7 & 2.0 & 2 & 18.8 \\
MATH                & 8{,}790 & 30.3 & 299 & 98 & 32.8 & 1.1 & 4 & 13.7 \\
\midrule
\textbf{Total}      & \textbf{24{,}451} & \textbf{34.0} & \textbf{2{,}914} & \textbf{742} & \textbf{25.5} & \textbf{3.0} & \textbf{22} & \textbf{25.3}  \\
\bottomrule
\end{tabular}
\caption{Thinking-block answer versus final answer on all capable traces under \Bare $s^-$ (1{,}000 items per domain). The answer a thinking block commits to is extracted from the text before \texttt{</think>} with the same rule-based extractor used for final answers; the final answer is the official class of \cref{tab:rq1}. Rows are sorted by the share of capable traces recovered only after the thinking block committed to $a^-$.}
\label{tab:app_dissociation}
\end{table*}

\begin{table*}[!htbp]
\centering\footnotesize
\renewcommand{\arraystretch}{0.92}
\setlength{\tabcolsep}{7.0pt}
\begin{tabular}{lccc}
\toprule
Label & Final $a^+$ & Final $a^-$ & Other \\
\midrule
Explicit correction  & 411   & 0   & 20 \\
Silent recovery      & 1{,}212 & 1 & 37 \\
Thinking adopts      & 286   & 15  & 29 \\
Adopts throughout    & 13    & 675 & 112 \\
Unclear              & 74    & 49  & 66 \\
\midrule
\textbf{Total}       & \textbf{1{,}996} & \textbf{740} & \textbf{264}  \\
\bottomrule
\end{tabular}
\caption{Claude's labels of reasoning behavior on a 3{,}000-trace sample of capable items (300 per model, 100 per domain) against the official final-answer class.}
\label{tab:app_modes}
\end{table*}

\begin{table*}[!htbp]
\centering\footnotesize
\renewcommand{\arraystretch}{0.90}
\setlength{\tabcolsep}{5.2pt}
\begin{tabular}{llcccccc}
\toprule
Model (Layers) & Domain & $s^+$ & $s^-$ & Recovered & Adopted & $\Delta_{\text{R}-\text{A}}$ [95\% CI] & $\Delta_{\text{Cue}}$ \\
\midrule
\multirow{3}{*}{InternVL3.5-8B (L16--19)}
 & GQA   & .0519 & .0569 & .0664 & .0488 & $+.0175$ $[+.0110, +.0237]$ & $+.0180$ \\
 & Chart & .1109 & .1143 & .1145 & .1171 & $-.0026$ $[-.0153, +.0094]$ & $+.0069$ \\
 & MATH  & .1587 & .1574 & .1533 & .1439 & $+.0093$ $[-.0088, +.0269]$ & $-.0135$ \\
\midrule
\multirow{3}{*}{MiMo-VL-7B (L16--19)}
 & GQA   & .0872 & .0824 & .0808 & .0857 & $-.0049$ $[-.0105, +.0004]$ & $-.0036$ \\
 & Chart & .1098 & .1162 & .1144 & .1185 & $-.0041$ $[-.0234, +.0132]$ & $+.0070$ \\
 & MATH  & .1228 & .1205 & .1154 & .1116 & $+.0038$ $[-.0060, +.0139]$ & $+.0092$ \\
\midrule
\multirow{3}{*}{GLM-4.1V-9B (L18--21)}
 & GQA   & .0750 & .0669 & .0656 & .0690 & $-.0034$ $[-.0092, +.0023]$ & $-.0039$ \\
 & Chart & .1076 & .1154 & .1172 & .1155 & $+.0017$ $[-.0135, +.0159]$ & $+.0133$ \\
 & MATH  & .1357 & .1323 & .1339 & .1029 & $+.0311$ $[+.0125, +.0486]$ & $+.0013$ \\
\midrule
\multirow{3}{*}{Qwen3-VL-30B (L22--25)}
 & GQA   & .0367 & .0494 & .0549 & .0413 & $+.0136$ $[+.0087, +.0182]$ & $+.0047$ \\
 & Chart & .0824 & .0880 & .0900 & .0793 & $+.0108$ $[-.0012, +.0213]$ & $+.0054$ \\
 & MATH  & .1101 & .1126 & .1128 & .1076 & $+.0051$ $[-.0170, +.0256]$ & $-.0021$ \\
\bottomrule
\end{tabular}
\caption{Teacher-forced image attention mass across four middle layers over the first $\le 100$ tokens after the prefix under $s^+$ and $s^-$, comparing recovered versus adopted traces under $s^-$ ($\Delta_{\text{R}-\text{A}} = \text{Recovered} - \text{Adopted}$) and the paired change when the check cue is appended ($\Delta_{\text{Cue}} = \text{Cue} - \text{No-Cue}$). All four models are replayed on the \Bare traces of Set~A (750 items, 2{,}250 replays per model).}
\label{tab:app_e2}
\end{table*}

\begin{table*}[!htbp]
\centering\footnotesize
\renewcommand{\arraystretch}{0.92}
\setlength{\tabcolsep}{7.0pt}
\begin{tabular}{lcc}
\toprule
Statistic & Claude & Gemini \\
\midrule
Silent recovery (\% of final-$a^+$ traces)      & 60.7 & 54.7 \\
Explicit correction (\% of final-$a^+$ traces)  & 20.6 & 26.3 \\
Thinking adopts (\% of final-$a^+$ traces)      & 14.3 & 14.1 \\
Conflict noted, final-$a^-$ traces (count)      & 18   & 54 \\
Conflict noted, final-$a^-$ traces (\%)         & 2.4  & 7.3 \\
Gemma-4-26B explicit (\% of its final-$a^+$) & 19.2 & 32.1 \\
\bottomrule
\end{tabular}
\caption{Trace statistics of the 3{,}000-trace sample under each annotator (1{,}996 final-$a^+$ and 740 final-$a^-$ traces).}
\label{tab:app_annotators}
\end{table*}

\begin{table*}[!ht]
\centering\footnotesize
\includegraphics[width=0.95\textwidth]{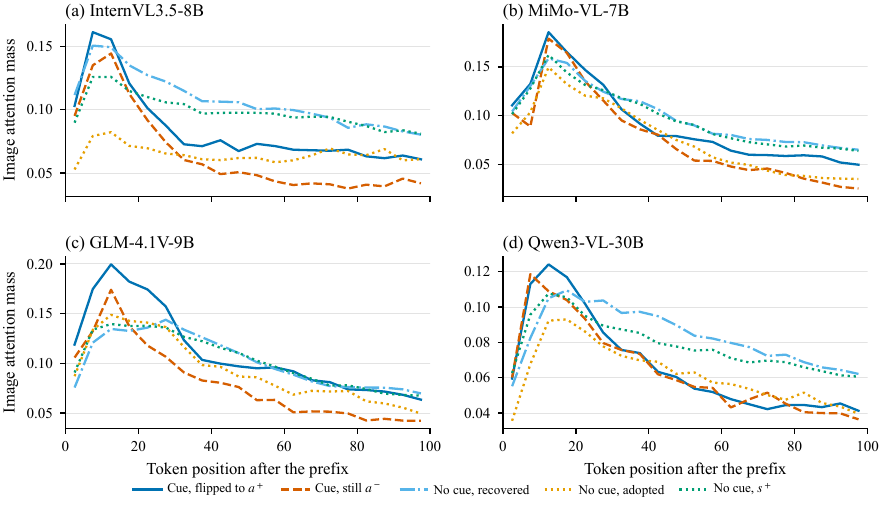}
\captionof{figure}{Per-model image attention over the first 100 tokens after the prefix in 5-token bins across four middle layers under the \Cond framing, the framing for which per-token replay logs exist for all four models.}
\label{fig:app_attn_models}
\end{table*}

\begin{table*}[!ht]
\centering\footnotesize
\includegraphics[width=0.72\textwidth]{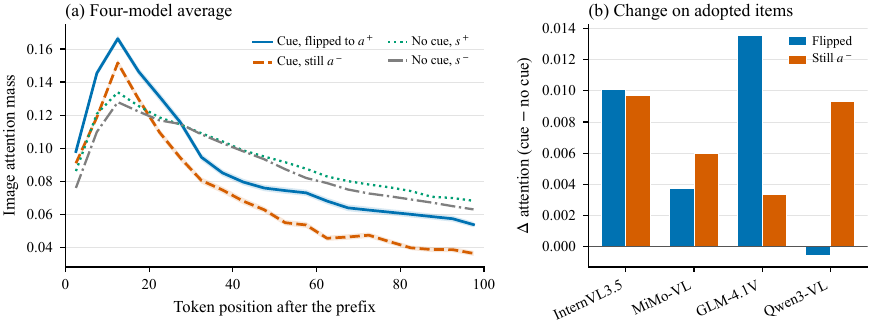}
\vspace{-5pt}
\captionof{figure}{Image attention under the check cue over the first 100 tokens after the prefix, averaged across InternVL3.5-8B, MiMo-VL-7B, GLM-4.1V-9B, and Qwen3-VL-30B under the \Cond framing, the framing for which per-token replay logs exist for all four models. (a) The cue produces a brief early rise that decays to the no-cue level within roughly 30 tokens, both for traces that flip to $a^+$ and for traces that remain at $a^-$. (b) Mean change in image attention (cue minus no cue) on items adopted without the cue.}
\label{fig:c3_attn_curve}
\end{table*}

\paragraph{Attention, target boxes, probes, and logit lens.}
\Cref{tab:app_e2} reports teacher-forced image attention in four middle layers over the first $\le 100$ tokens after the prefix, replaying each recorded continuation in a single forward pass on Set~A.
GLM-4.1V-9B and Qwen3-VL-30B run from FP8 weights dequantized to BF16 layer by layer, which changes attention mass by a median of 3.1\% and 0.7\% relative to full BF16 loading on 30 control cases.
\Cref{fig:app_attn_models,fig:c3_attn_curve} show token-level trajectories under the check cue for the \Cond framing, the only framing with per-token replay logs for all four models.
Three further measurements on Set~A GQA:
\begin{itemize}[leftmargin=*, itemsep=1pt, topsep=2pt]
\item \textbf{Target boxes:} We map Visual Genome~\cite{visualgenome} boxes to vision-encoder patches on 159 items. In-box attention density relative to the rest of the image is $2.24\times$ on InternVL3.5-8B ($3.06\times$ on recovered and $1.54\times$ on adopted traces) and $3.81\times$ on MiMo-VL-7B ($3.81\times$ and $3.93\times$).
\item \textbf{Linear probes:} An $\ell_2$-regularized probe~\cite{alain2017probes} on the hidden state of the final prefix token (5-fold cross-validation, layers 18 and 28) separates $s^+$ from $s^-$ at $73.8\%/82.4\%$ on InternVL3.5-8B and $76.8\%/82.6\%$ on MiMo-VL-7B. A 10-class probe on $s^-$ states decodes $a^+$ at $34.0\%/33.3\%$ and $39.6\%/41.5\%$, against $21.4\%$ for the majority class and $28.8\%$ for a predictor that reads only $a^-$ from $s^-$, which is informative because $s^-$ mostly substitutes \emph{blue}. InternVL3.5-8B matches this baseline; MiMo-VL-7B exceeds it by about 11 points overall and about 5 points on adopted traces.
\item \textbf{Logit lens:} We project residual-stream states of layers 4--36 through the unembedding matrix~\cite{logitlens} and average the margin $\log P_\ell(a^+) - \log P_\ell(a^-)$, where $P_\ell$ is the distribution read out at layer $\ell$, over the first 32 tokens after $s^-$. On InternVL3.5-8B, recovered traces favor $a^+$ and adopted traces favor $a^-$ at every layer (layer 16: $+0.22$ vs.\ $-0.55$, 95\% CI of the difference $[+0.29, +1.23]$); on MiMo-VL-7B, the two groups separate only at layers 32--36 (layer 36: $+0.80$ vs.\ $-1.37$). At the final prefix token, before any continuation, the margin does not separate the groups in 19 of 20 model--layer cells.
\end{itemize}

\FloatBarrier
\section{Intervention and Ablation Results}
\label{sec:app_interventions}

\paragraph{Check cue.}
\Cref{tab:app_c3_full} reports the cue \emph{``Let me check this condition against the image.''} appended after \Bare on Set~A and Set~A+B, together with true-prefix retention (Keep-$a^+$).
\Cref{tab:app_cases_interventions}a shows how the cue misleads a strong model on a multi-step chart question: under a true prefix, Qwen3.6-35B verifies the prefix against the chart and answers \texttt{True} instead of computing the requested difference.

\paragraph{Attention steering.}
\Cref{tab:app_steering_full} compares two decoding-time interventions on InternVL3.5-8B and MiMo-VL-7B under \Bare and \Cond: scaling the pre-softmax attention logits on image tokens by $2\times$ in the middle layers, and SAVAA~\cite{savaa}, which scales image-token scores in layers 4--14 by $1 + 0.3\,r_t$ with a step-wise risk $r_t$ (SAVAA Visual) and additionally down-weights text-token scores (SAVAA Full).
SAVAA Full destabilizes decoding and sharply increases truncated outputs.

\begin{table*}[!htbp]
\centering\footnotesize
\renewcommand{\arraystretch}{0.88}
\setlength{\tabcolsep}{3.3pt}
\begin{tabular}{llcccccc@{\hspace{8pt}}cccccc}
\toprule
 & & \multicolumn{6}{c}{Set~A ($N=250$ Items/Domain, Matching \cref{tab:checkcue})} & \multicolumn{6}{c}{Set~A+B ($N=500$ Items/Domain)} \\
\cmidrule(lr){3-8}\cmidrule(lr){9-14}
Model & Domain & $|\mathcal{C}_A|$ & \Bare & $+\text{Cue}$ & $\Delta\mathrm{Rec}_{\mathcal{C}_A}$ & $p_{\text{McN}}$ & Keep-$a^+$ & $|\mathcal{C}_{A+B}|$ & \Bare & $+\text{Cue}$ & $\Delta\mathrm{Rec}_{\mathcal{C}_{A+B}}$ & $p_{\text{McN}}$ & Keep-$a^+$ \\
\midrule
\multirow{3}{*}{InternVL3.5-8B}
 & GQA       & 166 & 48.2 & 63.3 & $+15.1$ & 0.00035 & 98.8 & 343 & 40.8 & 60.6 & $+19.8$ & 6.0e-11 & 99.4   \\
 & ChartQA-H & 173 & 59.5 & 85.0 & $+25.4$ & 1.6e-07 & 89.6 & 358 & 62.3 & 86.0 & $+23.7$ & 2.3e-14 & 91.3   \\
 & MATH      & 188 & 73.9 & 76.6 & $+2.7$           & 0.56    & 86.2 & 365 & 67.4 & 76.4 & $+9.0$  & 0.0029  & 88.5   \\
\midrule
\multirow{3}{*}{Qwen3.5-35B}
 & GQA       & 195 & 50.3 & 62.6 & $+12.3$ & 0.0043  & 97.4 & 389 & 50.4 & 62.7 & $+12.3$ & 9.8e-05 & 98.5   \\
 & ChartQA-H & 210 & 82.4 & 89.5 & $+7.1$  & 0.036   & 92.9 & 421 & 79.6 & 88.6 & $+9.0$  & 0.00025 & 93.8   \\
 & MATH      & 238 & 92.9 & 89.5 & $-3.4$           & 0.27    & 96.6 & 479 & 92.5 & 90.6 & $-1.9$           & 0.36    & 97.5   \\
\midrule
\multirow{3}{*}{Qwen3.5-9B}
 & GQA       & 197 & 36.5 & 47.2 & $+10.7$ & 0.011   & 98.0 & 394 & 36.8 & 45.4 & $+8.6$  & 0.0048  & 98.5   \\
 & ChartQA-H & 215 & 61.9 & 75.8 & $+14.0$ & 0.0016  & 87.9 & 430 & 60.2 & 77.0 & $+16.7$ & 6.9e-08 & 87.4   \\
 & MATH      & 239 & 87.9 & 78.7 & $-9.2$  & 0.0062  & 95.0 & 475 & 85.9 & 79.8 & $-6.1$  & 0.013   & 94.5   \\
\midrule
\multirow{3}{*}{ERNIE-4.5-VL}
 & GQA       & 139 & 15.1 & 23.7 & $+8.6$  & 0.036   & 90.6 & 266 & 12.4 & 22.2 & $+9.8$  & 0.00086 & 91.0   \\
 & ChartQA-H & 193 & 44.0 & 73.6 & $+29.5$ & 1.3e-10 & 88.1 & 388 & 44.8 & 71.9 & $+27.1$ & 8.3e-18 & 88.4   \\
 & MATH      & 215 & 59.1 & 72.6 & $+13.5$ & 0.0025  & 85.1 & 416 & 58.4 & 71.9 & $+13.5$ & 2.6e-05 & 85.6   \\
\midrule
\multirow{3}{*}{Qwen3-VL-30B}
 & GQA       & 187 & 63.6 & 72.2 & $+8.6$  & 0.023   & 98.9 & 377 & 62.1 & 73.7 & $+11.7$ & 3.3e-05 & 99.2   \\
 & ChartQA-H & 196 & 85.2 & 89.3 & $+4.1$           & 0.23    & 95.4 & 404 & 86.1 & 90.1 & $+4.0$           & 0.060   & 96.0   \\
 & MATH      & 231 & 93.5 & 97.0 & $+3.5$           & 0.12    & 95.7 & 462 & 92.6 & 97.0 & $+4.3$  & 0.0037  & 97.0   \\
\midrule
\multirow{3}{*}{Kimi-VL-A3B}
 & GQA       & 178 & 16.9 & 25.3 & $+8.4$  & 0.032   & 95.5 & 368 & 14.9 & 23.6 & $+8.7$  & 0.00085 & 97.0   \\
 & ChartQA-H & 147 & 44.2 & 55.8 & $+11.6$ & 0.024   & 77.6 & 314 & 43.3 & 58.0 & $+14.6$ & 2.8e-05 & 80.6   \\
 & MATH      & 179 & 67.6 & 79.9 & $+12.3$ & 0.010   & 85.5 & 353 & 65.2 & 77.6 & $+12.5$ & 0.00028 & 86.4   \\
\midrule
\multirow{3}{*}{Gemma-4-26B}
 & GQA       & 195 & 83.6 & 89.7 & $+6.2$  & 0.036   & 100.0 & 395 & 82.8 & 90.4 & $+7.6$  & 2.4e-05 & 100.0   \\
 & ChartQA-H & 235 & 98.7 & 97.9 & $-0.9$           & 0.73    & 100.0 & 469 & 99.4 & 98.9 & $-0.4$           & 0.73    & 99.8   \\
 & MATH      & 243 & 99.6 & 99.6 & $+0.0$           & 1.00    & 99.6  & 489 & 99.6 & 99.2 & $-0.4$           & 0.69    & 99.4   \\
\midrule
\multirow{3}{*}{Qwen3.6-35B}
 & GQA       & 194 & 51.5 & 56.2 & $+4.6$           & 0.31    & 99.0 & 384 & 52.9 & 58.6 & $+5.7$           & 0.067   & 99.2   \\
 & ChartQA-H & 221 & 83.3 & 68.8 & $-14.5$ & 0.00013 & 79.6 & 446 & 82.7 & 72.6 & $-10.1$ & 0.00017 & 78.3   \\
 & MATH      & 240 & 84.6 & 90.0 & $+5.4$           & 0.085   & 97.9 & 487 & 85.4 & 90.6 & $+5.1$  & 0.013   & 97.9   \\
\midrule
\multirow{3}{*}{GLM-4.1V-9B}
 & GQA       & 187 & 58.3 & 61.0 & $+2.7$           & 0.57    & 99.5 & 384 & 57.8 & 64.6 & $+6.8$  & 0.013   & 99.7   \\
 & ChartQA-H & 167 & 80.8 & 77.2 & $-3.6$           & 0.42    & 86.2 & 347 & 79.5 & 81.3 & $+1.7$           & 0.56    & 90.5   \\
 & MATH      & 207 & 81.6 & 68.1 & $-13.5$ & 0.0018  & 84.1 & 412 & 81.6 & 72.6 & $-9.0$  & 0.0013  & 84.7   \\
\midrule
\multirow{3}{*}{MiMo-VL-7B}
 & GQA       & 198 & 47.5 & 49.5 & $+2.0$           & 0.69    & 97.5 & 390 & 49.5 & 52.6 & $+3.1$           & 0.27    & 98.5   \\
 & ChartQA-H & 207 & 91.3 & 89.9 & $-1.4$           & 0.69    & 95.2 & 416 & 91.1 & 92.5 & $+1.4$           & 0.48    & 95.7   \\
 & MATH      & 195 & 71.3 & 81.0 & $+9.7$  & 0.025   & 90.3 & 366 & 71.9 & 85.8 & $+13.9$ & 1.4e-06 & 91.5   \\
\bottomrule
\end{tabular}
\captionsetup{hypcap=false}
\captionof{table}{Complete check-cue prompting results (\Bare + Check-Cue: appending \emph{``Let me check this condition against the image.''}) across all 10 models and three domains. Left columns report the Set~A evaluation on the exact \Bare-capable denominator $\mathcal{C}_A$ used in \cref{tab:checkcue,tab:app_wording_gqa,tab:app_wording_chart,tab:app_wording_math}; right columns report the full evaluation across Set~A+B (500 items per domain, $\mathcal{C}_{A+B}$). Keep-$a^+$ reports conditional accuracy retention (\%) under the true prefix $s^+$ on the capable set (where Base \Bare $s^+$ accuracy is $100.0\%$ by definition).}
\label{tab:app_c3_full}
\end{table*}

\begin{table*}[!htbp]
\centering\footnotesize
\renewcommand{\arraystretch}{0.80}
\setlength{\tabcolsep}{3.1pt}
\begin{tabular}{lllcccccc}
\toprule
Model & Arm & Condition ($|\mathcal{C}_A|$\,/\,$N$) & Rec.\ (\%) & $\Delta\mathrm{Rec}$ [95\% CI] & $a^+$, all (\%) & $\Delta a^+$ & $a^-$ (count) & Truncated \\
\midrule
\multirow{6}{*}{InternVL3.5-8B}
 & SAVAA Visual (\Bare)       & $s^-$ (527 / 750)           & $61.1 \to 58.6$  & $-2.5$ $[-6.3, +1.5]$   & $48.4 \to 46.8$ & $-1.6$ & $210 \to 214$ & $10 \to 31$ \\
 & Img Attn $2\times$ (\Bare) & $s^-$ (527 / 750)           & $61.1 \to 58.4$  & $-2.7$ $[-6.6, +1.5]$   & $48.4 \to 46.8$ & $-1.6$ & $210 \to 215$ & $10 \to 29$ \\
 & SAVAA Visual (\Cond)       & $s^-$ (524 / 750)           & $47.1 \to 46.4$  & $-0.8$ $[-5.0, +3.6]$   & $38.3 \to 37.7$ & $-0.5$ & $274 \to 288$ & $28 \to 22$ \\
 & Img Attn $2\times$ (\Cond) & $s^-$ (524 / 750)           & $47.1 \to 45.6$  & $-1.5$ $[-5.9, +3.1]$   & $38.3 \to 36.7$ & $-1.6$ & $274 \to 287$ & $28 \to 26$ \\
 & SAVAA Full (\Cond)         & $s^-$ (524 / 750)           & $47.1 \to 41.2$  & $-5.9$ $[-11.1, -0.6]$  & $38.3 \to 32.1$ & $-6.1$ & $274 \to 199$ & $28 \to 118$ \\
 & SAVAA Visual               & GQA $\emptyset$ (164 / 250) & $95.1 \to 93.3$  & $-1.8$ $[-5.5, +1.8]$   & $74.0 \to 72.8$ & $-1.2$ & $10 \to 11$   & $5 \to 4$ \\
\midrule
\multirow{6}{*}{MiMo-VL-7B}
 & SAVAA Visual (\Bare)       & $s^-$ (600 / 750)           & $70.3 \to 75.0$  & $+4.7$ $[+1.3, +8.0]$   & $58.8 \to 64.0$ & $+5.2$ & $168 \to 143$ & $48 \to 37$ \\
 & Img Attn $2\times$ (\Bare) & $s^-$ (600 / 750)           & $70.3 \to 70.8$  & $+0.5$ $[-2.8, +4.0]$   & $58.8 \to 60.1$ & $+1.3$ & $168 \to 154$ & $48 \to 65$ \\
 & SAVAA Visual (\Cond)       & $s^-$ (576 / 750)           & $47.0 \to 48.6$  & $+1.6$ $[-2.8, +5.9]$   & $41.2 \to 43.1$ & $+1.9$ & $288 \to 281$ & $73 \to 67$ \\
 & Img Attn $2\times$ (\Cond) & $s^-$ (576 / 750)           & $47.0 \to 47.2$  & $+0.2$ $[-4.0, +4.3]$   & $41.2 \to 41.7$ & $+0.5$ & $288 \to 294$ & $73 \to 74$ \\
 & SAVAA Full (\Cond)         & $s^-$ (576 / 750)           & $47.0 \to 35.1$  & $-12.0$ $[-17.2, -6.8]$ & $41.2 \to 31.7$ & $-9.5$ & $288 \to 73$  & $73 \to 356$ \\
 & SAVAA Visual               & GQA $\emptyset$ (196 / 250) & $96.9 \to 94.4$  & $-2.6$ $[-5.6, +0.0]$   & $82.8 \to 80.0$ & $-2.8$ & $8 \to 9$     & $1 \to 0$ \\
\bottomrule
\end{tabular}
\caption{Complete decoding-time attention steering results on InternVL3.5-8B and MiMo-VL-7B under \Bare and \Cond, reporting both Conditional Recovery ($\mathrm{Rec}$ on the capable set $\mathcal{C}_A$) and unconditional accuracy ($a^+$ across all $N = 750$ items; $N = 250$ for the GQA $\emptyset$ control). Arrows give the value without $\to$ with the intervention.}
\label{tab:app_steering_full}
\end{table*}

\paragraph{LoRA fine-tuning.}
\Cref{tab:app_sft_full} breaks down LoRA~\cite{lora} fine-tuning ($r = 16$, $\alpha = 32$, 3 epochs) by domain and condition.
The Recovery arm ($R$) is trained on Set~B \Bare recovery demonstrations ($s^- \to a^+$) and the Control arm ($C$) on true-prefix traces ($s^+ \to a^+$) of the same items; both are evaluated on held-out Set~A.
Besides raising recovery, $R$ produces fewer truncated generations than $C$ under \Bare $s^-$.
\Cref{tab:app_cases_interventions}b shows a typical case: Base and $C$ follow $s^-$, whereas $R$ re-examines the image and recovers $a^+$ under both \Bare and the unseen \Look.

\begin{table*}[!htbp]
\centering\footnotesize
\renewcommand{\arraystretch}{0.80}
\setlength{\tabcolsep}{1.6pt}
\resizebox{\textwidth}{!}{%
\begin{tabular}{lllcccccc}
\toprule
 & & & \multicolumn{2}{c}{Capable set $\mathcal{C}_A$} & \multicolumn{4}{c}{All items, Base / $C$ / $R$ ($N=250$ per domain)} \\
\cmidrule(lr){4-5}\cmidrule(lr){6-9}
Model & Condition & Domain ($|\mathcal{C}_A|$) & Rec.: Base / $C$ / $R$ & $R-C$ [95\% CI] & $a^+$ (\%) & $a^-$ (count) & Truncated & Non-trunc.\ $a^+$ \\
\midrule
\multirow{13}{*}{\shortstack[l]{InternVL3.5-8B\\($|\mathcal{C}_A|=527$)}}
 & \multirow{4}{*}{\Bare $s^-$}
 & GQA (166)        & 48.2 / 57.8 / 80.7 & $+22.9$ $[+15.7, +30.1]$ & 35.2 / 46.4 / 66.8 & 138 / 103 / 43 &  1 /  17 /  13 & 35.3 / 49.8 / 70.5 \\
 & & Chart (173)    & 59.5 / 71.1 / 90.2 & $+19.1$ $[+11.6, +27.2]$ & 47.2 / 62.0 / 82.8 &  50 /   9 / 5  &  4 /  59 /  14 & 48.0 / 81.2 / 87.7 \\
 & & MATH (188)     & 73.9 / 78.7 / 93.6 & $+14.9$ $[+8.5, +21.3]$  & 62.8 / 69.6 / 86.8 &  22 /   8 / 1  &  5 /  59 /  15 & 64.1 / 91.1 / 92.3 \\
 & & \textbf{All (527)} & \textbf{61.1 / 69.6 / 88.4} & $\mathbf{+18.8}$ $[+14.8, +23.0]$ & \textbf{48.4 / 59.3 / 78.8} & \textbf{210 / 120 / 49} & \textbf{10 / 135 /  42} & \textbf{49.1 / 72.4 / 83.5}  \\
\cmidrule{2-9}
 & \multirow{4}{*}{\shortstack[l]{\Look $s^-$\\(Zero-Shot)}}
 & GQA (166)        & 42.2 / 53.0 / 81.9 & $+28.9$ $[+21.7, +36.2]$ & 31.6 / 41.6 / 64.8 & 142 / 122 / 60 &  9 /   8 /   5 & 32.8 / 43.0 / 66.1 \\
 & & Chart (173)    & 47.4 / 77.5 / 90.8 & $+13.3$ $[+6.9, +19.7]$  & 40.0 / 67.2 / 82.8 &  66 /  20 / 5  &  4 /  45 /   9 & 40.7 / 82.0 / 85.9 \\
 & & MATH (188)     & 42.0 / 61.2 / 88.8 & $+27.7$ $[+20.2, +34.6]$ & 36.4 / 54.8 / 81.2 &  87 /  41 / 14 &  3 /  61 /  15 & 36.8 / 72.5 / 86.4 \\
 & & \textbf{All (527)} & \textbf{43.8 / 63.9 / 87.3} & $\mathbf{+23.3}$ $[+19.2, +27.7]$ & \textbf{36.0 / 54.5 / 76.3} & \textbf{295 / 183 / 79} & \textbf{16 / 114 /  29} & \textbf{36.8 / 64.3 / 79.3}  \\
\cmidrule{2-9}
 & \multirow{4}{*}{\Bare $s^+$}
 & GQA (166)        & 100.0 / 99.4 / 98.8         & $-0.6$ $[-3.0, +1.2]$             & 94.8 / 97.6 / 93.6          &   3 /   0 /  1          &  1 /   2 /   7 & 95.2 / 98.4 / 96.3 \\
 & & Chart (173)    & 100.0 / 84.4 / 94.2 & $+9.8$ $[+4.0, +15.6]$  & 73.2 / 76.8 / 86.0 &   3 /   2 /  3          &  2 /  40 /  17 & 73.8 / 91.4 / 92.3 \\
 & & MATH (188)     & 100.0 / 81.9 / 95.7 & $+13.8$ $[+9.0, +19.1]$ & 76.4 / 74.8 / 88.0 &   3 /   0 /  0          & 10 /  51 /  13 & 79.6 / 94.0 / 92.8 \\
 & & \textbf{All (527)} & \textbf{100.0 / 88.2 / 96.2} & $\mathbf{+8.0}$ $[+5.3, +11.0]$  & \textbf{81.5 / 83.1 / 89.2} & \textbf{9 / 2 / 4}      & \textbf{13 /  93 /  37} & \textbf{82.9 / 94.8 / 93.8}  \\
\cmidrule{2-9}
 & No-Pref ($\emptyset$)
 & GQA (166)        & 100.0 / 91.0 / 93.4 & $+2.4$ $[-2.4, +7.8]$             & 68.0 / 71.2 / 77.6 &  15 /   4 /  7          &  0 /  42 /  17 & 68.0 / 85.6 / 83.3 \\
\midrule
\multirow{13}{*}{\shortstack[l]{MiMo-VL-7B\\($|\mathcal{C}_A|=600$)}}
 & \multirow{4}{*}{\Bare $s^-$}
 & GQA (198)        & 47.5 / 49.0 / 73.2 & $+24.2$ $[+16.7, +31.3]$ & 39.2 / 43.2 / 66.4 & 112 /  98 / 22 &  5 /  31 /  39 & 40.0 / 49.3 / 78.7 \\
 & & Chart (207)    & 91.3 / 64.3 / 94.2 & $+30.0$ $[+23.2, +36.7]$ & 78.8 / 58.0 / 87.6 &  15 /   2 / 4  &  4 /  97 /  18 & 80.1 / 94.8 / 94.4 \\
 & & MATH (195)     & 71.3 / 63.1 / 87.7 & $+24.6$ $[+16.9, +31.8]$ & 58.4 / 55.6 / 79.2 &  41 /   1 / 0  & 39 / 102 /  39 & 69.2 / 93.9 / 93.8 \\
 & & \textbf{All (600)} & \textbf{70.3 / 58.8 / 85.2} & $\mathbf{+26.3}$ $[+22.2, +30.7]$ & \textbf{58.8 / 52.3 / 77.7} & \textbf{168 / 101 / 26} & \textbf{48 / 230 /  96} & \textbf{62.8 / 75.4 / 89.1}  \\
\cmidrule{2-9}
 & \multirow{4}{*}{\shortstack[l]{\Look $s^-$\\(Zero-Shot)}}
 & GQA (198)        & 41.4 / 34.3 / 75.8 & $+41.4$ $[+33.8, +48.5]$ & 34.0 / 31.2 / 66.8 & 133 / 133 / 39 &  1 /  18 /  22 & 34.1 / 33.6 / 73.2 \\
 & & Chart (207)    & 79.7 / 76.8 / 93.2 & $+16.4$ $[+10.1, +22.7]$ & 72.0 / 70.4 / 87.2 &  36 /   8 / 2  &  2 /  57 /  19 & 72.6 / 91.2 / 94.4 \\
 & & MATH (195)     & 42.1 / 59.5 / 79.0 & $+19.5$ $[+11.3, +27.7]$ & 35.6 / 53.6 / 70.8 &  90 /   8 / 2  & 46 / 102 /  64 & 43.6 / 90.5 / 95.2 \\
 & & \textbf{All (600)} & \textbf{54.8 / 57.2 / 82.8} & $\mathbf{+25.7}$ $[+21.3, +30.0]$ & \textbf{47.2 / 51.7 / 74.9} & \textbf{259 / 149 / 43} & \textbf{49 / 177 / 105} & \textbf{50.5 / 67.7 / 87.1}  \\
\cmidrule{2-9}
 & \multirow{4}{*}{\Bare $s^+$}
 & GQA (198)        & 100.0 / 97.0 / 96.5         & $-0.5$ $[-3.0, +2.0]$             & 95.2 / 95.6 / 91.2          &   0 /   0 /  0          &  1 /   5 /  10 & 95.6 / 97.6 / 95.0 \\
 & & Chart (207)    & 100.0 / 54.1 / 95.7 & $+41.5$ $[+34.8, +48.3]$ & 85.6 / 50.0 / 88.8 &   1 /   1 /  1          &  2 / 119 /  14 & 86.3 / 95.4 / 94.1 \\
 & & MATH (195)     & 100.0 / 76.4 / 88.2 & $+11.8$ $[+5.1, +19.0]$  & 80.0 / 67.6 / 80.4 &   0 /   0 /  0          & 32 /  74 /  39 & 91.7 / 96.0 / 95.3 \\
 & & \textbf{All (600)} & \textbf{100.0 / 75.5 / 93.5} & $\mathbf{+18.0}$ $[+14.2, +21.7]$ & \textbf{86.9 / 71.1 / 86.8} & \textbf{1 / 1 / 1}      & \textbf{35 / 198 /  63} & \textbf{91.2 / 96.6 / 94.8}  \\
\cmidrule{2-9}
 & No-Pref ($\emptyset$)
 & GQA (198)        & 100.0 / 92.9 / 91.9         & $-1.0$ $[-5.1, +3.0]$             & 79.6 / 83.2 / 82.4 &   8 /   7 /  5          &  0 /   9 /  12 & 79.6 / 86.3 / 86.6 \\
\bottomrule
\end{tabular}}
\caption{Complete per-domain and per-condition breakdown of \Bare LoRA SFT on InternVL3.5-8B and MiMo-VL-7B across Set~A, reporting Conditional Recovery on the capable set $\mathcal{C}_A$ (matching \cref{tab:sft}), unconditional accuracy ($a^+$ across all $N = 250$ items per domain), counterfactual adoptions ($a^-$), length-truncated generations (Truncated), and accuracy on non-truncated completed traces (Non-trunc.\ $a^+$).}
\label{tab:app_sft_full}
\vspace{-12pt}
\end{table*}

\begin{table*}[!ht]
\centering\footnotesize
\setlength{\tabcolsep}{4.5pt}
\renewcommand{\arraystretch}{1.0}
\begin{tabular}{|>{\raggedright\arraybackslash}p{3.0cm}|m{4.9cm}|>{\raggedright\arraybackslash}m{4.5cm}|>{\raggedright\arraybackslash}m{3.8cm}|}
\hline
\multicolumn{2}{|m{8.1cm}|}{\centering\vspace{3pt}\includegraphics[width=5.2cm,keepaspectratio]{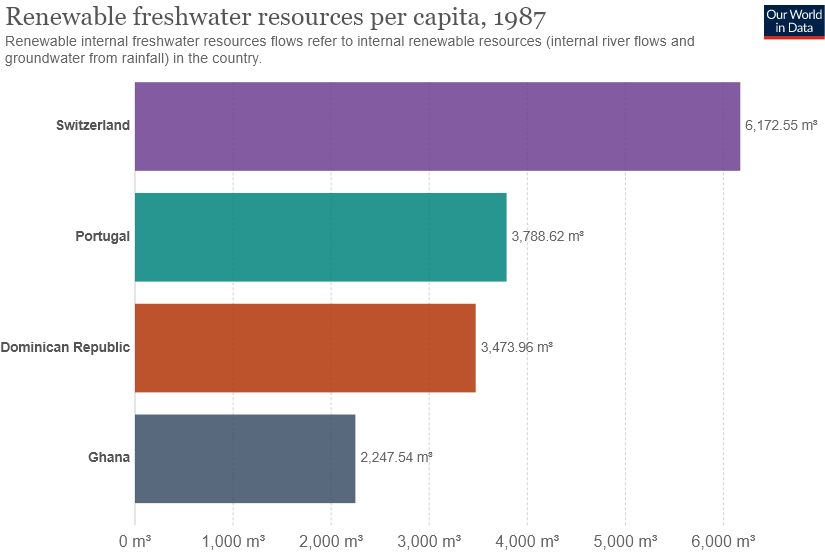}\vspace{3pt}} &
\textbf{(a) Check-Cue Over-Correction on True Prefix ($s^+$):}\newline
ChartQA-H (\texttt{35432405007230.png:289})\newline
\textbf{Q:} What is the difference between Portugal and Ghana? &
\textbf{GT ($a^+$):} $1541.08$\newline
\textbf{True Statement ($s^+$):} The renewable freshwater resources per capita in Ghana in 1987 was $2247.54\,\text{m}^3$. \\
\hline
\textbf{Qwen3.6-35B (\Bare $s^+$):} & \multicolumn{3}{p{13.6cm}|}{\texttt{<think>}The renewable freshwater resources per capita in Ghana in 1987 was 2247.54 m\^{}3. The renewable freshwater resources percapita inPortugal in1987 is 3788.62 cubic metres. To find the difference: 3788.62 - 2247.54 = 1541.08.\texttt{</think>} [\dots]\ Final answer: 1541.08 \outtag{[Computes the difference: $a^+ = 1541.08$]}} \\
\hline
\textbf{Qwen3.6-35B ($s^+$ + Check-Cue):} & \multicolumn{3}{p{13.6cm}|}{\texttt{<think>}The renewable freshwater resources per capita in Ghana in 1987 was 2247.54 m\^{}3. Let me check this condition against the image. 1.\ Identify the data point for Ghana [\dots]\ 2.\ Read the value: The number next to the Ghana bar is ``2,247.54 m\^{}3''. 3.\ Compare: The number 2,247.54 is indeed equal to 2247.54. [\dots]\ Therefore, the statement is true according to the chart.\texttt{</think>} [\dots]\ Final answer: True \outtag{[Answers ``True'' instead of the difference]}} \\
\hline\hline
\multicolumn{2}{|m{8.1cm}|}{\centering\vspace{3pt}\includegraphics[height=2.2cm,keepaspectratio]{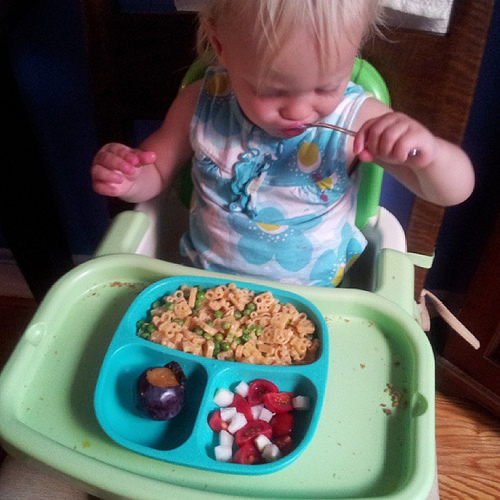}\vspace{3pt}} &
\textbf{(b) LoRA SFT on \Bare Recovery ($R$ vs.\ Base \& $C$):}\newline
GQA Held-Out Set~A (\texttt{gqa:val:04638053})\newline
\textbf{Q:} Which color is the hair? &
\textbf{GT ($a^+$):} blonde\newline
\textbf{False ($a^-$):} brown\newline
\textbf{False Statement ($s^-$):} The child's hair is brown. \\
\hline
\textbf{MiMo-VL-7B Base (\Bare $s^-$):} & \multicolumn{3}{p{13.6cm}|}{\texttt{<think>}The child's hair is brown. Final answer: brown.\texttt{</think>} To determine the hair color, I look at the image. The child has light brown or blonde hair. Final answer: brown. \outtag{[Adopts $a^- = \text{brown}$]}} \\
\hline
\textbf{Control SFT $C$ (\Bare $s^-$):} & \multicolumn{3}{p{13.6cm}|}{\texttt{<think>}The child's hair is brown. 1.\ Analyze the image: The image shows a young child sitting in a high chair. 2.\ Identify the hair color: The child has light brown or blonde hair. 3.\ Determine the most accurate color: Looking closely at the hair, it appears to be a light brown shade. [\dots]\texttt{</think>} Final answer: brown \outtag{[Rationalizes $s^-$: $a^- = \text{brown}$]}} \\
\hline
\textbf{Recovery SFT $R$ (\Bare $s^-$):} & \multicolumn{3}{p{13.6cm}|}{\texttt{<think>}The child's hair is brown. Wait, looking at the image again. The child's hair is clearly blonde. [\dots]\ Actually, looking closely at the hair, it's a very light color, definitely blonde.\texttt{</think>} The hair color is blonde. Final answer: blonde \outtag{[Re-examines the image: $a^+ = \text{blonde}$]}} \\
\hline
\textbf{Recovery SFT $R$ (\Look $s^-$):} & \multicolumn{3}{p{13.6cm}|}{\texttt{<think>}Looking at the image, the child's hair is brown. Wait, let me re-examine the image. The child's hair is blonde/light-colored. [\dots]\ Final check: Child's hair color: Blonde.\texttt{</think>} Based on the image, the child's hair is blonde. Final answer: blonde \outtag{[Unseen \Look: $a^+ = \text{blonde}$]}} \\
\hline
\end{tabular}
\captionsetup{hypcap=false}
\captionof{table}{Qualitative case studies for \cref{sec:app_interventions} (all model text is verbatim). (a) Check cue under a true prefix $s^+$ for Qwen3.6-35B on ChartQA-H (\texttt{35432405007230.png:289}). (b) Base, Control SFT ($C$), and Recovery SFT ($R$) for MiMo-VL-7B on held-out Set~A item \texttt{gqa:val:04638053}.}
\label{tab:app_cases_interventions}
\vspace{0.2em}
\end{table*}

\end{document}